\documentclass[journal]{IEEEtran}
\usepackage{amsmath,amsfonts}
\usepackage{array}
\usepackage[caption=false,font=normalsize,labelfont=sf,textfont=sf]{subfig}
\usepackage{textcomp}
\usepackage{stfloats}
\usepackage{url}
\usepackage{verbatim}
\usepackage{graphicx}
\usepackage{algorithm}\usepackage{algpseudocode}
\usepackage{amssymb}
\usepackage{caption} 
\usepackage{multirow}
\usepackage{graphicx} 

\def\BibTeX{{\rm B\kern-.05em{\sc i\kern-.025em b}\kern-.08em
    T\kern-.1667em\lower.7ex\hbox{E}\kern-.125emX}}
\usepackage{balance}
\usepackage{caption}
\usepackage{enumitem}

\begin{document}

\title{SHFL: Secure Hierarchical Federated Learning Framework for Edge Networks}

\author{

\IEEEauthorblockN{ Omid Tavallaie, \textit{Member, IEEE}, Kanchana Thilakarathna, Suranga Seneviratne, Aruna Seneviratne, \textit{Senior Member, IEEE}, and Albert Y. Zomaya, \textit{Fellow, IEEE}}
\thanks{\textit{Omid Tavallaie, Kanchana Thilakarathna, Suranga Seneviratne, and Albert Y. Zomaya are with the School of Computer Science, The University of Sydney, Sydney, NSW 2006, Australia. Emails: omid.tavallaie@sydney.edu.au; kanchana.thilakarathna@sydney.edu.au; suranga.seneviratne@sydney.edu.au; albert.zomaya@sydney.edu.au.} 

\textit{Aruna Seneviratne is with the School of Electrical Engineering and Telecommunications, University of New South Wales (UNSW), Sydney,
NSW 2052, Australia. E-mail: a.seneviratne@unsw.edu.au.}
}
}
\maketitle

\begin{abstract}
Federated Learning (FL) is a distributed machine learning paradigm designed for privacy-sensitive applications that run on resource-constrained devices with non-Identically and Independently Distributed (IID) data. Traditional FL frameworks adopt the client-server model with a single-level aggregation (AGR) process, where the server builds the global model by aggregating all trained local models received from client devices. However, this conventional approach encounters challenges, including susceptibility to model/data poisoning attacks. In recent years, advancements in the Internet of Things (IoT) and edge computing have enabled the development of hierarchical FL systems with a two-level AGR process running at edge and cloud servers. In this paper, we propose a Secure Hierarchical FL (SHFL) framework to address poisoning attacks in hierarchical edge networks. By aggregating trained models at the edge, SHFL employs two novel methods to address model/data poisoning attacks in the presence of client adversaries: 1) a client selection algorithm running at the edge for choosing IoT devices to participate in training, and 2) a model AGR method designed based on convex optimization theory to reduce the impact of edge models from networks with adversaries in the process of computing the global model (at the cloud level). The evaluation results reveal that compared to state-of-the-art methods, SHFL significantly increases the maximum accuracy achieved by the global model in the presence of client adversaries applying model/data poisoning attacks. 
\end{abstract}


\section{Introduction}

Federated Learning (FL) \cite{fedavg, pyramid, secure-ccs22, secureagg} is a distributed client-server machine learning framework designed for privacy-aware applications when non-Identically and Independently Distributed (IID) data is continuously generated on client (mobile) devices. By training a model based on the local dataset, each client device sends an update to the server to compute a global model in the model aggregation (AGR) process. Compared to centralized learning, where model training is entirely handled by the server, federated learning significantly reduces server-side computational costs by outsourcing and parallelizing the training process. FL provides strong privacy guarantees for client devices as the training data never leaves the client's device. Federated learning systems face several practical challenges: 1) mobile devices have only sporadic access to power and network connectivity, 2) the set of users participating in each training round is changing unpredictably, 3) the client device continuously generates training data based on the usage of the device by a user, which could be significantly different from the distribution of generated data by other users (non-IID). Due to the substantial overhead of training models on a large number of client devices, only a small number of clients are selected in a round of training based on a client selection algorithm running on the server \cite{pyramid}.

 \begin{figure}[t]
 \includegraphics[width= 65 mm, height=50 mm]{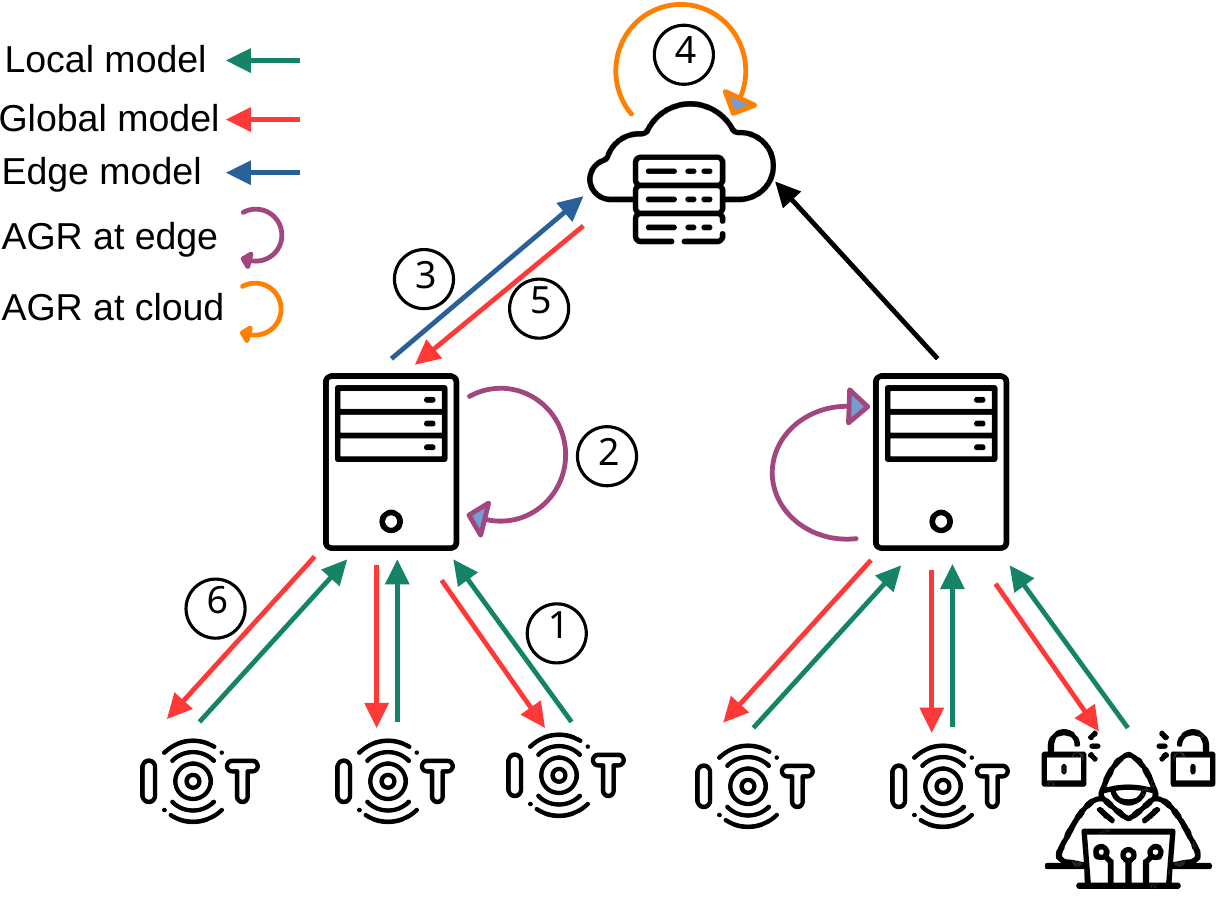} \centering
  \caption{Model/data poisoning attack in hierarchical FL where model AGR is performed at two levels (edge and cloud).\vspace{-3 mm}}
  \label{fig:attack}
\end{figure}

\begin{figure*}[ht!] 
  \centering 
  \subfloat[\textnormal{Conventional FL}]{ 
    \includegraphics[width=65 mm, height=45 mm]{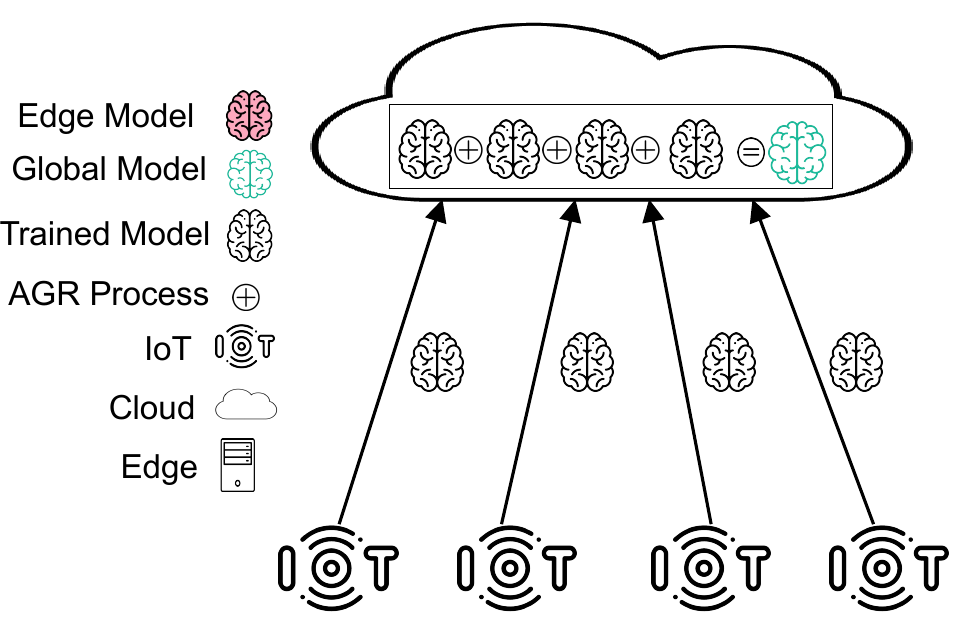} 
  }
  \hspace{10 mm} 
  \subfloat[\textnormal{Hierarchical FL}]{ 
    \includegraphics[width=47 mm, height=45 mm]{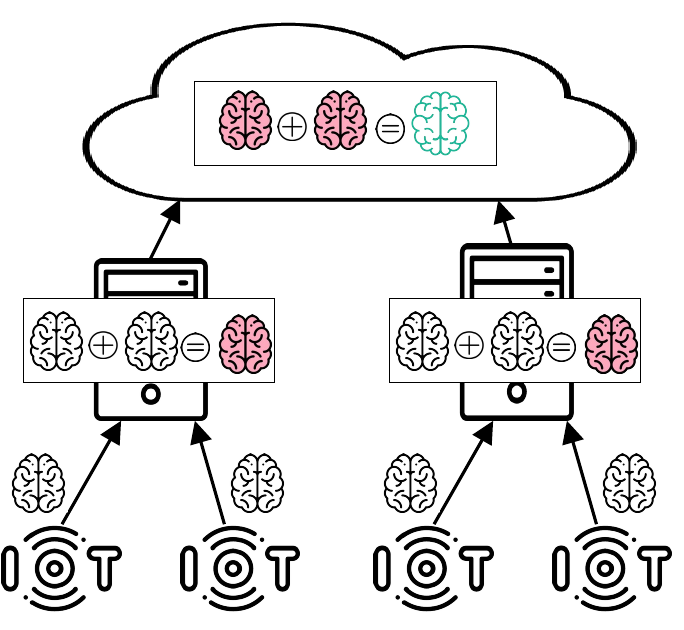} 
  }
  \caption{Comparing the AGR process in conventional and hierarchical FL. }
  \label{fig:attack2}
\end{figure*}

In conventional FL frameworks, a single aggregator is used at the cloud server that receives updates (trained models) from client devices and performs model AGR to build a global model, which is returned to client devices for the next round of training. This type of FL frameworks face technical challenges such as inefficient communication and system failures. For instance, the single cloud aggregator can become a bottleneck when many clients send updates simultaneously. Hierarchical Federated Learning (HFL) \cite{acmj} is an extension of the conventional FL framework which is designed to address the specific challenges and opportunities in edge computing environments \cite{multitask, smartgrid, hier, itifs23}. As Fig. \ref{fig:attack} shows, a hierarchical FL framework consists of three main levels: 1) Internet of Things (IoT) devices that generate data and perform model training, 2) edge servers that receive trained models from selected IoT devices, aggregate them, and send the aggregated model to the cloud server, and 3) the cloud server that receives the aggregated models from edge servers and performs model AGR to create the global model. Fig. \ref{fig:attack2} compares the AGR process in conventional and hierarchical FL frameworks. FL is vulnerable to Byzantine attacks when attackers upload poisoned local model updates \cite{ijcai} to impact the AGR process and corrupt the global model. This attack can be performed by manipulating the local model gradients sent from client devices to the aggregator \cite{usenix20} (Fig. \ref{fig:attack}). While several methods have been proposed in recent years to address model poisoning attacks \cite{model2, tdsc1, tdsc2, tdsc3, tdsc4, tdsc5} for conventional FL, investigating the impact of these attacks on hierarchical edge networks (with two levels of AGR) has been left as an unresolved research problem.

In this paper, we propose a secure hierarchical FL framework called SHFL to address model/data poisoning attacks on edge networks by mitigating the impact of poisoned updates on the AGR process running at two levels of the computational hierarchy, edge and cloud servers. The main contributions of this paper are summarized as follows:
\begin{itemize}[left=0 pt]
\item \textit{Defense against model/data poisoning attacks}: By proposing a client selection algorithm at the edge level and a new AGR process at the cloud level, SHFL mitigates the impact of poisoned models on the process of building the global model. Evaluation results revealed that, compared to state-of-the-art methods (FedAvg \cite{fedavg}, Multi-krum \cite{multi-krum}, and Trimmed-mean \cite{trimmed-mean}) SHFL reduces the impact of the poisoning attack on the accuracy by up to 31\%.

\item \textit{Client selection at the edge:} SHFL filters out potentially poisoned models by comparing the gradients of local models with the global model by using $L_{2}$-norm distance. Clients with updates closest to the global model are ranked, and the least trustworthy (based on the distance) are excluded. After filtering out an estimated number of attackers, SHFL selects a random subset of the most trusted clients to participate in training. This process ensures that benign updates are prioritized, improving the robustness of the edge-aggregated model.

\item \textit{Optimizing the AGR process}: At the cloud level, SHFL leverages convex optimization theory to combine edge server models while mitigating the impact of adversarial updates. By computing the $L_{2}$-norm distance between each edge model and the global model and by considering the size of the data processed by each edge, it computes the optimal weight assigned to each edge model. Trustworthy models with smaller distances (to the global model) and larger data contributions are given higher weights in the AGR process. This ensures that the AGR process prioritizes reliable updates while minimizing the influence of potentially malicious models. 

\item \textit{Fidelity}: To assess SHFL's performance in both adversarial and non-adversarial settings, we conducted comprehensive sets of experiments by using three image classification datasets (MNIST, FMNIST, and CIFAR-10) across 28 scenarios, considering both IID and non-IID data distributions, and employing various neural network models. Even in the absence of attacks, SHFL can be applied to enhance the global model's accuracy, demonstrating its benefits beyond just defending against adversarial threats.

\end{itemize}

The remainder of this paper is organized as follows: section II provides background knowledge about model/data poisoning attacks on FL and defense mechanisms. Section III reviews the related work and state-of-the-art methods. Section IV details our proposed method, SHFL. Section V explains the attack models. Section VI evaluates the performance of SHFL and state-of-the-art methods against model/data poisoning attacks. Finally, section VII concludes this paper. Table {\ref{table:1}} presents the main symbols used in this paper.

\begin{table}[hb]
\renewcommand{\arraystretch}{1.2}
\caption{Main symbols used in this paper.}
\label{table:1}
\centering
\begin{tabular}{|*{40}{p{10 mm}|p{68 mm}|}}
\hline
\textbf{\hspace{-1 mm}Symbol}  &\textbf{Definition}\\ \hline \hline
$CS_{edge}$  &Set of all edge servers\\ \hline
$CS_{j}$  &Set of all IoT nodes connected to the edge server $j$\\ \hline
$TS_{j}$  &Set of IoT nodes selected for training in edge network $j$\\ \hline
$T_{i}^{k}$  &Trust metric of node $i$ at round $k$\\ \hline
$b_{i}^{k}$  & $L_{2}$-norm distance between $AM_{i}^{k}$ and $AM_{global}^{k}$\\ \hline
$D_{i}^{k}$  &Total amount of data used for training at edge network $i$\\ \hline
$M_{i}^{k}$  &Trained model of node $i$ at round $k$\\ \hline
$d_{i}^{k}$ &The amount of training data used by node $i$ at round $k$\\  \hline
$c_{i}^{k}$ &Weight of the node $i's$ model in the AGR at the edge\\  \hline
$AM_{i}^{k}$  &Aggregated model of edge $i$ at round $k$\\  \hline
$w_{i}^{k}$  &Weight of edge $i$ at round $k$ in the cloud AGR\\  \hline
${\|i-j\|}_{2}$ &$L_{2}$-norm distance between models $i$ and $j$\\  \hline
$AM_{global}^{k}$   &Global model computed at round $k$\\  \hline
$u_{i}$   &Utility function of node $i$\\  \hline
$\tau$ &Total weight for the AGR running at the cloud level\\  \hline
$\zeta$ &Minimum weight of an edge model in the AGR process\\  \hline
\end{tabular} 
\end{table}

\section{Background and Problem Statement}
Before introducing our proposed method, in this section, we briefly explain model/data poisoning attacks on FL and the defense mechanisms that are used against them.\vspace{-3 mm}
\subsection{Poisoning Attacks}
Poisoning attacks in FL refer to malicious attacks aim to compromise the integrity of the distributed learning process. The main types of attacks in FL are data poisoning \cite{iotj223, data1, data2, data3, lf} and model poisoning \cite{NDSS21, model1, sp22}. A data poisoning attack involves planting poisoned data into the training dataset with the aim of changing the model’s weights to suit the attacker’s objectives. Data poisoning attacks are categorized as either targeted or untargeted. The attacker can inject data samples that are slightly manipulated to cause minimal effect or inject a large number of biased samples to cause a significant impact on the model. Untargeted model poisoning attacks seek to reduce the accuracy of the global model across all test inputs. This type of attack can affect a large number of FL clients and may remain undetected for an extended period \cite{sp22}. In model poisoning attacks, adversaries try to manipulate the model itself rather than the training data by changing the local model gradients sent from client devices to the aggregator. In targeted model poisoning attacks, the attacker's objective is to cause the global model to misclassify specific test examples according to the attacker's target class. By using this approach, an adversary can implant covert backdoors into the global model, causing images containing a trojan trigger to be classified as attacker-specified labels (referred to as the backdoor attack \cite{back1, back2}). In this paper, we focus on untargeted model/data poisoning attacks.
\vspace{-4 mm}
\subsection{Defense Mechanisms Against Untargeted Poisoning Attacks}

The literature has explored numerous strategies to enhance the robustness of traditional FL against Byzantine or compromised clients. The primary approach involves replacing the standard average AGR rule of FL with a more resilient one, commonly referred to as a defense. In non-adversarial FL settings, FedAvg \cite{fedavg} is an effective AGR rule. Federated Averaging is the only AGR rule implemented at large scale (e.g., by Apple and Google) for FL applications. Below, we outline the various types of robust AGR rules designed to protect FL from untargeted poisoning attacks:

\begin{itemize}[left= 0 pt]
\item \textit{Dimension-wise filtering} independently screens for potentially malicious values in each dimension of clients' updates. In the context of neural networks, a filter or kernel is typically a small matrix that is slid over the input data to perform operations like feature extraction. In dimension-wise filtering, the filter is applied independently to each channel or dimension of the input tensor. Trimmed-mean \cite{trimmed-mean} is a method that aggregates each dimension of input updates separately. It sorts the values of the $jth$ dimension of all updates. It then removes the $m$ largest and smallest values for that dimension and computes the average of the remaining values as the aggregate for dimension $j$.

\item \textit{Vector-wise filtering} aims to eliminate potentially poisoned client updates in their entirety. Unlike dimension-wise filtering, vector-wise filtering assesses the entire update from a client. If an update is deemed suspicious or malicious based on certain criteria, the whole update is discarded. This approach helps ensure that no part of a compromised update influences the model. By focusing on entire updates, vector-wise filtering can be particularly effective in scenarios where malicious clients attempt to disrupt the learning process with highly coordinated and sophisticated attacks.

\item \textit{Norm-bounding} mitigates the effect of poisoned updates by scaling their norms. This approach uses a robust AGR process that enforces a fixed threshold on the $L_{2}$-norm of all client updates. Effective poisoned updates generally exhibit high norms, which can disproportionately influence the global model. Additionally, norm-bounding serves as a proactive defense against covert but powerful poisoning attempts, ensuring that the influence of all updates stays within a controlled and secure range.
\end{itemize}

\section{Related Work}
FL-WBC was introduced by \cite{nips21} as a client-defense mechanism to address model poisoning attacks that have already compromised the global model. The core concept of FL-WBC is to find the parameter space where persistent attack effects reside and disrupt that space during training. FedClean \cite{fedclean} is built on the principles of active learning. It incorporates a reputation mechanism that monitors each agent's credibility and helps in selecting trustworthy agents for model training, along with an update quality control mechanism that detects malicious updates. The optimal scenario for FedClean is to filter out any malicious updates before they are incorporated into model training. The authors of FedClean assumed that some agents might submit malicious updates in a given round but aim to contribute honestly in future rounds to earn rewards.

The authors of \cite{usenix20} assumed that the attacker has control over certain client devices and modifies the local model parameters that these devices send to the aggregator. The attacker may or may not be aware of the AGR rule used by the aggregator. The authors demonstrated that this attack cannot be detected by the well-known defense mechanisms that attempt to identify malicious data based on its negative impact on the error rate of the trained model. The findings in this paper suggest that this attack can greatly increase the error rates of models trained using federated learning.

SPPFL \cite{pst21} is a scalable privacy-preserving FL framework designed to ensure data confidentiality for users during the FL training process in a multi-server environment. Privacy threats are mainly due to the servers, which might attempt to infer users' private details from the information provided by users. This paper assumes that the servers are honest-but-curious. SPPFL defends against poisoning attacks, including label flipping and backdoor attacks, while safeguarding users' private information. To identify and remove outlier updates from the AGR process, the pairwise distances between gradients are calculated, and a set of sufficiently close gradients is selected. Shamir's secret sharing and Lagrange interpolation are employed by the servers to prevent the extraction of useful information from the updates.

In \cite{NDSS21}, the authors propose a novel robust AGR algorithm, called divide-and-conquer (DnC), to defend against model poisoning attacks, inspired by defenses mechanisms against data poisoning in centralized learning systems. Spectral analysis techniques, like singular value decomposition, are used by these defense mechanisms to detect and remove outliers in poisoned data. The concept behind DnC is that a model poisoning update has an impact only when it deviates substantially from benign updates. DnC initially calculates the principal component, that is, the direction of maximum variance, from the set of input updates. Afterward, it computes the scalar products between the updates and the principal component which is called projections. As a final step, DnC filters out a constant fraction of updates with the largest projections. However, it is computationally impossible to carry out spectral analysis on extremely high-dimensional model updates in FL. Thus, by performing dimensionality reduction, DnC enables spectral analysis of the input updates and ensures effective detection of malicious updates.

\cite{trimmed-mean} introduced the Trimmed-mean AGR to enhance the robustness of FL. This approach provides a simple yet effective way to mitigate the risk of poisoning attacks by sorting clients' updates for each model parameter and trimming a predefined percentage of the most smallest and largest values. The mean of the remaining values is then computed to form the aggregated update. The trimmed mean leverages the statistical properties of the updates, making it a robust defense mechanism without needing to identify which clients are malicious.

\cite{multi-krum} Multi-krum is an AGR method operates by selecting updates from multiple clients and evaluating their Euclidean distances to other updates. For each update, the algorithm computes the sum of distances to its closest neighbors and selects the updates with the smallest sums to find the models which are more consistent with the majority of updates. This method effectively filters out poisoned updates when an attacker generates malicious updates that have larger distances from benign ones. By focusing on updates that are most similar to others, Multi-krum ensures that the aggregated update is derived from the most reliable and consistent updates.

\cite{icdcs21} proposed BAFFLE, a backdoor detection mechanism for FL frameworks. BAFFLE uses data of multiple clients for detecting model poisoning. Through a round-based feedback loop, BAFFLE exploits the diversity of clients' data for validating the global model to decide whether a model update is genuine or not. FLCert \cite{itifs22} guarantees the security of FL against model poisoning for a bounded number of malicious clients. This method makes the groups of clients for creating different global model to testify a test input. Two methods are proposed in \cite{itifs22} for creating client's group: 1) random selection, and 2) selection based on model similarities.

To evaluate other related work, in \cite{sp22}, authors discussed the gap between academic studies and practical implementation. This paper also proposed a model poisoning attack called PGA where adversary can modify the updates of compromised clients directly. By using the Stochastic Gradient Ascent (SGA) algorithm, this attack increases the loss on some benign data to generate a malicious model. Then, it fine-tunes the model weights by normalizing the poisoned model in the range of $L_{2}$-norms of benign updates. In this works, authors assumed some attackers generate benign updates that are used in normalizing the weights of the poisoned model.

To the best of our knowledge, SHFL is the first work to address model/data poisoning attacks in hierarchical edge networks with two level AGR process. To ensure SHFL filters poisoned update at two levels, we propose two novel methods for the client selection at the edge and the model AGR at the cloud. In the computational hierarchy, each IoT node is connected directly to an edge server, where SHFL first filters poisoned updates by assessing the $L_{2}$-norm distance of a model update to the global model (computed at the cloud). Then, we use a novel AGR method to minimize the influence of edge models linked to malicious nodes in the AGR process running at the cloud level for building the global model.
\vspace{-3 mm}
\section{SHFL: Secure Hierarchical FL Framework}
In this section, we explain our proposed method, SHFL, in detail and demonstrate how it computes the optimal weight of an edge model in the cloud AGR process based on convex optimization theory.\vspace{-3 mm}
\subsection{Designing the Trust Metric}
Each IoT node $i\in CS_{j}$ connected to the edge server $j$ sends a model update $M_{i}$ at each round to its corresponding edge server. The edge server performs the first level of AGR, computing the aggregated model $AM_{j}$ by selecting a subset of clients. To facilitate the client selection process, we define a trust metric at the edge level, which determines the reliability of each client's model update by comparing its similarity to the global model ($AM_{global}^{k}$). The trust metric is calculated as the $L_{2}$-norm distance between the local model $M_{i}$ of a client and the global model $AM_{global}^{k}$. This metric is updated dynamically at every round as the global model evolves. The trust metric is used to rank all the nodes connected to edge server $j$, helping to prioritize clients with updates that are more aligned with the global model (likely to be benign). The trust metric for node $i$ at round $k$ is computed as follows:
\begin{equation}\label{eq:c1}
t_{i}^{k}={\|M_{i}^{k}-AM_{global}^{k}\|}_{2}.
\end{equation}
where ${\|\|}_{2}$ represents the $L_{2}$-norm, which measures the Euclidean distance between the two models.
\vspace{-3 mm}
\subsection{Client Selection at the edge}
Once the trust metric $t_{i}^{k}$ has been calculated for each IoT node connected to an edge server, the nodes are ranked based on their $L_{2}$-norm distances from the global model. Nodes with larger $L_{2}$-norm distances, which indicate greater divergence from the global model, are considered less trustworthy and are filtered out based on the estimated number of attackers per edge. This ensures that only the most reliable nodes remain for participation in the training process. After filtering, a subset of the remaining trusted nodes is selected for training according to the percentage of clients allowed to participate. Crucially, this final selection is done randomly to ensure fairness among the trusted clients. This random selection is especially beneficial in non-IID data settings, where each node may possess data with different distributions or labels. In such scenarios, selecting clients purely based on trust metrics could lead to over-representation of nodes with certain data characteristics, skewing the global model. Random selection ensures that all trusted nodes, regardless of slight differences in their trust metrics or data distributions, have an equal chance to contribute, which helps maintain a balanced and representative global model. For example, consider an edge server with 10 connected nodes, where up to 3 of them are estimated to be potential attackers. After calculating the trust metric, the 3 nodes with the largest $L_{2}$-norm distances are filtered out, leaving 7 trusted nodes. If 30\% of clients are to be selected for training, 3 nodes are randomly chosen from the 7 remaining clients that create $TS_{j}$ (for edge $j$). By introducing this random selection, SHFL enhances both fairness and robustness, ensuring that the global model is trained with diverse data while maintaining protection against adversarial attacks. Algorithm \ref{alg1} shows the process of selecting $m$ clients at edge $i$ when the estimated number of attackers connected to the edge is $a$.
\vspace{-3 mm}
\subsection{AGR at the Edge}
After creating the training set ($TS_{j}$), which includes the IoT devices participating in building the edge model $AM_{j}^{k}$, edge server asks these devices to start training. The total amount of data used by each IoT device for training is considered in the AGR process. This is crucial because, while the $L_{2}$-norm (used in the trust metric) captures the quality of the model updates by measuring how closely they align with the global model, it does not account for the quantity of data used for generating these updates. To ensure that devices contributing larger datasets have a proportionally greater influence on the aggregated model, we compute a weight $c_{i}^{k}$ for each IoT device $i \in TS_{j}$ (connected to edge $j$), based on the amount of data ($d_{i}^{k}$) used by the device for training at round $k$:
\vspace{-1 mm}
\begin{equation}\label{eq:weight}
c_{i}^{k} = \dfrac{d_{i}^{k}}{\sum_{v \in TS_{j}} d_{v}^{k}}.
\end{equation}
This ensures that nodes with more data have a larger impact on the edge AGR process, reflecting their greater contribution to the overall training. Using these computed weights, edge server $j$ builds the aggregated model $AM_{j}^{k}$ by taking a weighted sum of model (gradient) updates from all participating nodes:
\vspace{-1 mm}
\begin{equation}\label{eq:aggregate}
AM_{j}^{k} = \sum_{v \in TS_{j}} c_{v}^{k} \times M_{v}^{k}.
\end{equation}
By considering the trained model (via the trust metric based on $L_{2}$-norm) and data quantity (via $c_{i}^{k}$), SHFL ensures that the AGR process remains effective even in non-adversarial scenarios where larger datasets typically provide more reliable and generalizable model updates. Hence, giving greater weight to such updates improves the robustness and accuracy of the edge-aggregated model.

\begin{algorithm}[t]
\caption{SHFL's client selection process running at edge $i$ in round $k$ when the estimated number of attackers per edge network is $a$ and $m$ clients are selected for training.}
\begin{algorithmic}[1]
\For{$\forall i \in CS_{j}$}
    \State {Receive $M_{i}^{k}$ }
     \State { Compute $t_{i}^{k}={\|M_{i}^{k}-AM_{global}^{k}\|}_{2}$} 
\EndFor
\State Create the set $DS_{j}$ by sorting the values of $t_{j}^{k}(\forall j \in CS_{i})$
\State Remove $a$ elements with largest values of $t_{i}^{k}$ and create the set $ES_{j}$ for all remaining elements
\State Select $m$ clients randomly from the set $ES_{j}$ and save them in the set $TS_{j}$
\end{algorithmic}\label{alg1}
\end{algorithm}

\subsection{Designing an AGR  process by using Convex Optimization}
At the cloud level, SHFL aggregates the edge models to build the global model. Since an edge server may select an attacker in the AGR process, SHFL aims to detect and minimize the impact of the poisoned edge model on the AGR process by reducing its weight to a minimum value. To achieve this, SHFL considers the $L_{2}$-norm distance between an aggregated edge model $i$ at round $k$ and the global model, i.e.,
\begin{equation}\label{eq:aggregate}
b_{i}^{k} ={\|AM_{i}^{k}-AM_{global}^{k}\|}_{2},
\end{equation}
and the total amount of data ($D_{i}^{k}=\sum_{v \in TS_{i}}d_{v}^{k}$) used in the edge network $i$ for model training at IoT devices. To differentiate edge models, we define metric $x_{i}$ for edge $i$ in a way that by reducing the distance to the global model or increasing the amount of data used for training, the value is increased:
\begin{equation}\label{eq:7.2}
x_{i}^{k}= \dfrac{D_{i}^{k}}{min_{(j \in CS_{edge})}(D_{j}^{k})} \times  \dfrac{max_{(j \in CS_{edge})}(b_{j}^{k})}{b_{i}^{k}}.
\end{equation}
Here, it should be noted that $D_{i}^{k}$ and $b_{i}^{k}$ are always greater than zero. This formulation ensures that edge models using more training data or producing updates closer to the global model, are weighted more heavily, while those further from the global model are penalized. By incorporating $x_{i}^{k}$ and applying convex optimization theory, we address the problem of assigning optimal weights to edge models, even when some models may include attacker nodes in the AGR process. For each edge server $i \in CS_{edge}$, we define a utility function $u_{i}(w_{i}^{k})$ to represent the benefit that edge $i$ gains when $w_{i}^{k}$ is used as its assigned weight in the AGR process of round $k$. Utility functions, including exponential, reciprocal, and logarithmic, are commonly applied in the modeling of distributed systems, as they offer flexibility in representing diverse system behaviors \cite{iotdi, icdcs23, ipsn}. For the AGR problem, we define the utility function $u_{i}$ for edge $i$ using a logarithmic formulation:
\begin{equation}\label{eq:7.1}
u_{i}(w_{i}^{k})= x_{i}^{k}log(w_{i}^{k}+1).
\end{equation}
This function remains differentiable with respect to $w_{i}^{k}$ for every edge $i \in CS_{edge}$. Notably, it exhibits strict concavity, as its second partial derivative with respect to $w_{i}^{k}$ is negative, i.e., \vspace{-3 mm}

\begin{equation}\label{eq:4.8}
\frac{\partial ^{2} u_{i}(w_{i}^{k})}{\partial (w^{k}_{i})^{2}}=-\dfrac{x_{i}^{k}}{(1+w_{i}^{k})^2} < 0.
\end{equation}
We formulate the task of determining the optimal value of $w_{i}^{k}$ as an optimization problem that seeks to maximize the total sum of all utility function outputs, defined as:
 \vspace{-1 mm}
\begin{equation}\label{eq:4.12}
\begin{array}{c@{\qquad}c@{\qquad}c}
\vspace{2 mm}\text{maximize} \;\; \sum_{\forall i \in CS_{edge}}{u_{i}(w_{i}^{k})} \\
\vspace{2 mm}\text{subject to:}\\
w_{i}^{k} \ge \zeta, \\
\sum_{\forall i \in CS_{edge}}{w_{i}^{k}} \le \tau.
\end{array} 
\end{equation}
where $\zeta$ is the minimum weighting threshold and $\tau$ is a fixed value used in the weighted average AGR process by the cloud server as: \vspace{-1 mm}
\begin{equation}\label{eq:4.13}
AM^{k+1}_{global}=\sum_{i \in CS_{edge}}\dfrac{w_{i}^{k}}{\tau}M^{k}_{i}
\end{equation}
To solve the optimization problem \eqref{eq:4.12}, we apply the Lagrange multiplier method \cite{lag1, lag2}. This approach involves defining a Lagrange function $\mathcal{L}(w, v, \lambda_{i})$, which incorporates the constraints by subtracting them, scaled by the Lagrange multipliers $v$ and $\lambda_{i}$, from the objective function. The Lagrange function is expressed as:
\begin{equation}\label{eq:415}
\begin{aligned}
& \mathcal{L}(w,v,\lambda) = \\
&\sum_{\forall i \in CS_{edge}}{u_{i}(w_{i}^{k})} - v\left(\sum_{\forall i \in CS_{edge}}{w_{i}^{k}} - \tau\right) - \\ 
&\sum_{\forall i \in CS_{edge}}{\lambda_{i}(\zeta - w_{i}^{k})}.
\end{aligned}
\end{equation}

Based on the Karush-Kuhn-Tucker (KKT) optimality conditions, the solution of \eqref{eq:4.12} is determined by identifying the values of $w_{i}^{k}(\forall i \in CS_{j})$ that meet the following criteria:

\begin{enumerate}
\item{Primal feasibility conditions:\newline
$\zeta-w_{i}^{k} \le 0,\;\;\;\; (\sum_{\forall i \in CS_{edge}}{w_{i}^{k}})-\tau \le 0.$}\newline\vspace{-3 mm}
\item{Dual feasibility conditions:\newline
$v \ge 0,\;\;\;\;  \lambda_{i} \ge 0.$}\newline\vspace{-4 mm}
\item{Stationary condition:\newline
$\frac {\partial (\sum_{\forall i \in CS_{edge}}{u_{i}(w_{i}^{k})})}{\partial w_{i}^{k}}-\\v\frac {\partial(\sum_{\forall i \in CS_{edge}}{w_{i}^{k}}-\tau)}{\partial w_{i}^{k}}
-\lambda_{i}\frac {\partial(\zeta-w_{i}^{k})}{\partial w_{i}^{k}}=0.$}\newline\vspace{-3 mm}
\item{Complementary slackness conditions:\newline\vspace{3 mm}
$v(\sum_{\forall i \in CS_{edge}}{w_{i}^{k}}-\tau)=0  ,\;\;\;\;  \lambda_{i}(\zeta-w_{i}^{k})=0.$}\vspace{-3 mm}
\end{enumerate}
By determining the values for variables $v$ and $\lambda_{i}$ that satisfy these conditions, the optimal value of  $w_{i}^{k}$ for each edge $\forall i \in CS_{edge}$ is found by:\\
\begin{equation}\label{eq:15.3}
w_{i}^{k}=
\begin{cases}
   \zeta,& \;\;\;\;\text{if } i \in X,\\\\
    f^{k}_{i},& \;\;\;\;\text{otherwise,}           
\end{cases}
\end{equation}
where the set $X$ and  $f^{k}_{i}$ are defined as
\begin{equation}\label{eq:15.1}
\begin{gathered}
\hspace{-50 mm}X=\{i|i\in CS_{edge},\\
\zeta \ge \dfrac{x_{i}^{k}(\tau + |CS_{edge}|-1)-\sum_{j \in \{CS_{edge}- \{i\}\}}x_{j}^{k}}{ \sum_{\forall j\in CS_{edge}}x_{j}^{k} }\},
\end{gathered}
\end{equation}
\begin{equation}\label{eq:15.4}
f_{i}^{k} =\dfrac{x_{i}^{k}(\tau -(|X|\times\zeta)+ |V|-1)-\sum_{j \in \{V- \{i\}\}}x_{j}^{k}}{ \sum_{\forall j\in V}x_{j}^{k} },
\end{equation}
and the set $V$ is defined by
\begin{equation}\label{eq:15.2}
\begin{gathered}
\hspace{-50 mm}V=\{i|i\in CS_{edge},\\
\zeta < \dfrac{x_{i}^{k}(\tau + |CS_{edge}|-1)-\sum_{j \in \{CS_{edge}- \{i\}\}}x_{j}^{k}}{ \sum_{\forall j\in CS_{edge}}x_{j}^{k} }\}.
\end{gathered}
\end{equation}

\begin{algorithm}[b]
\caption{SHFL's model AGR process running at the cloud server at round $k$.}
\begin{algorithmic}[1]
\For{$\forall$ edge  $i \in CS_{edge}$}
    \State {Receive $AM_{i}^{k}, D_{i}^{k}$ }
     \State {Compute $b_{i}^{k}$ by \eqref{eq:aggregate} }
\EndFor
\For{$\forall$ edge  $i \in CS_{edge}$}
    \State {Compute $x_{i}^{k}$ by \eqref{eq:7.2}}
\EndFor

\State {Compute set $X$ by \eqref{eq:15.1}}
\State {Compute set $V$ by \eqref{eq:15.2}}
\For{$\forall$ edge  $i \in CS_{edge}$}
\If{$i \in X$} 
    \State $w_{i}^{k}=\zeta$ 
\Else 
    \State {$w_{i}^{k} =f_{i}^{k}$} (computed by \eqref{eq:15.4})
\EndIf
\EndFor
\State {$AM^{k+1}_{global}=\sum_{i \in CS_{edge}}\dfrac{w_{i}^{k}}{\tau}M^{k}_{i}$}

\end{algorithmic}\label{alg2}
\end{algorithm}

Algorithm \ref{alg2} shows the the AGR process running at the cloud server in round $k$ to generate the global model for round $k+1$. The values of $\zeta$ and $\tau$ should be chosen based on both the complexity of the data distribution and the severity of the attack model. For example, in the case of PGA attack, selecting an attacker during a training round can drastically reduce the accuracy of the global model, with potential long-lasting effects that may not be recoverable in subsequent rounds. In such scenarios, we set $\zeta = 0.1$ and $\tau = 10$ when the average weight of an edge model is normalized to 1 (i.e., 10 edge servers, each with a weight of $1$). On the other hand, for the LF attack, where selecting an attacker in a single round has a less severe impact on model accuracy, a higher value for $\zeta$ can be considered. This allows for greater tolerance in handling such attacks without significantly reducing model performance.
\vspace{-2 mm}
\section{The Attack Model}
After reviewing the most recent works on poisoning attacks in FL (\cite{tdsc1, tdsc2, tdsc3, tdsc4, tdsc5, icml1, cvf1, nips1}), we selected PGA (model poisoning) and LF (data poisoning) as the two most significant attacks for the evaluation of SHFL. These attacks have been extensively studied in recent years and have shown a higher impact on the global model compared to other methods. In this section, we briefly explain each attack and analyze their effects based on the number of attackers in the network.\vspace{-3 mm}
\subsection{Projected Gradient Ascent (PGA)}
The PGA attack \cite{sp22} (designed by Google), aims to increase the loss on benign data by using SGA during the training process. In this attack model, the attacker uses benign data to train the model by applying SGA, which modifies the gradients in a direction that maximizes the loss (by negating the loss value during gradient computation). After applying SGA, the PGA attack normalizes the generated poisoned update's norm according to the average norm of benign updates. This attack assumes that the attacker has access to benign models, which can occur in settings where some adversarial nodes act as benign clients to generate legitimate updates. However, this assumption may not hold in hierarchical edge networks, such as IoT systems, where devices are only connected to local edge servers via short-range wireless communications (e.g., Bluetooth or local WiFi). To address this, we adapt the PGA attack for our hierarchical framework (which consists of three levels of computation) and normalize the poisoned update by using the $L_{2}$-norm of the global model.\vspace{-3 mm}
\subsection{Label Flipping (LF)}
The LF attack is a well-known data poisoning method that can be challenging to detect. Unlike the PGA attack, which manipulates model updates directly, LF modifies only the training data by flipping the labels, with the goal of corrupting the model’s decision-making process. The attacker intentionally changes the labels of training data to mislead the training process, causing the model to learn an incorrect decision boundary. In our implementation of the LF attack, the attacker node changes each training sample's label to a randomly selected one to poison the training process. As in this paper we just focus on untargeted poisoning attacks, we did not consider scenarios when all the data for a specific label are changed to another label with the target of reducing the accuracy for a specific class.\vspace{-4 mm}
\begin{figure}[t]
  \centering
  \begin{minipage}{0.24\textwidth}
    \centering
    \includegraphics[width=43mm, height=38mm]{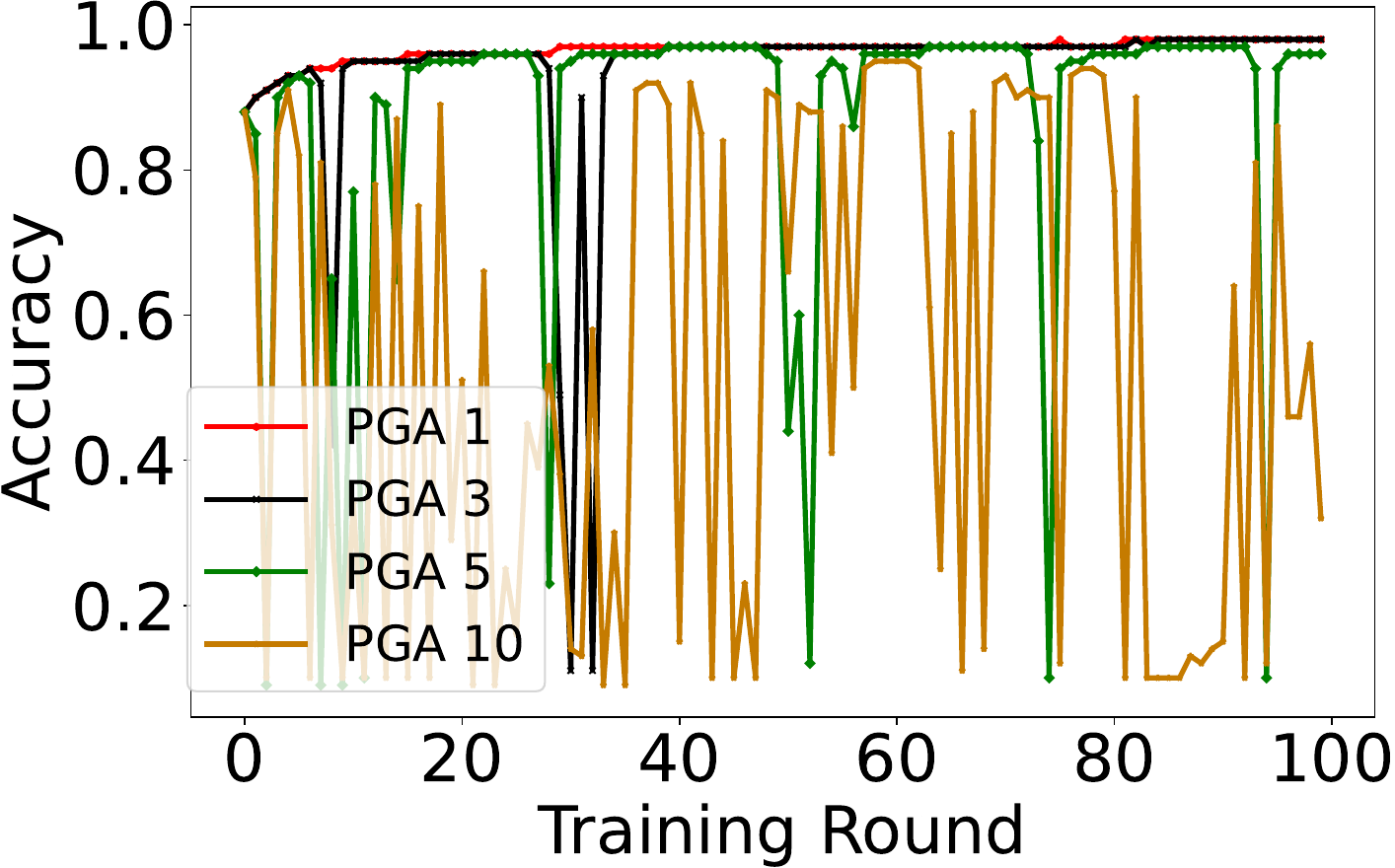}
    \caption*{(a) PGA}
  \end{minipage}
  \hfill
  \begin{minipage}{0.24\textwidth}
    \centering
    \includegraphics[width=43mm, height=38mm]{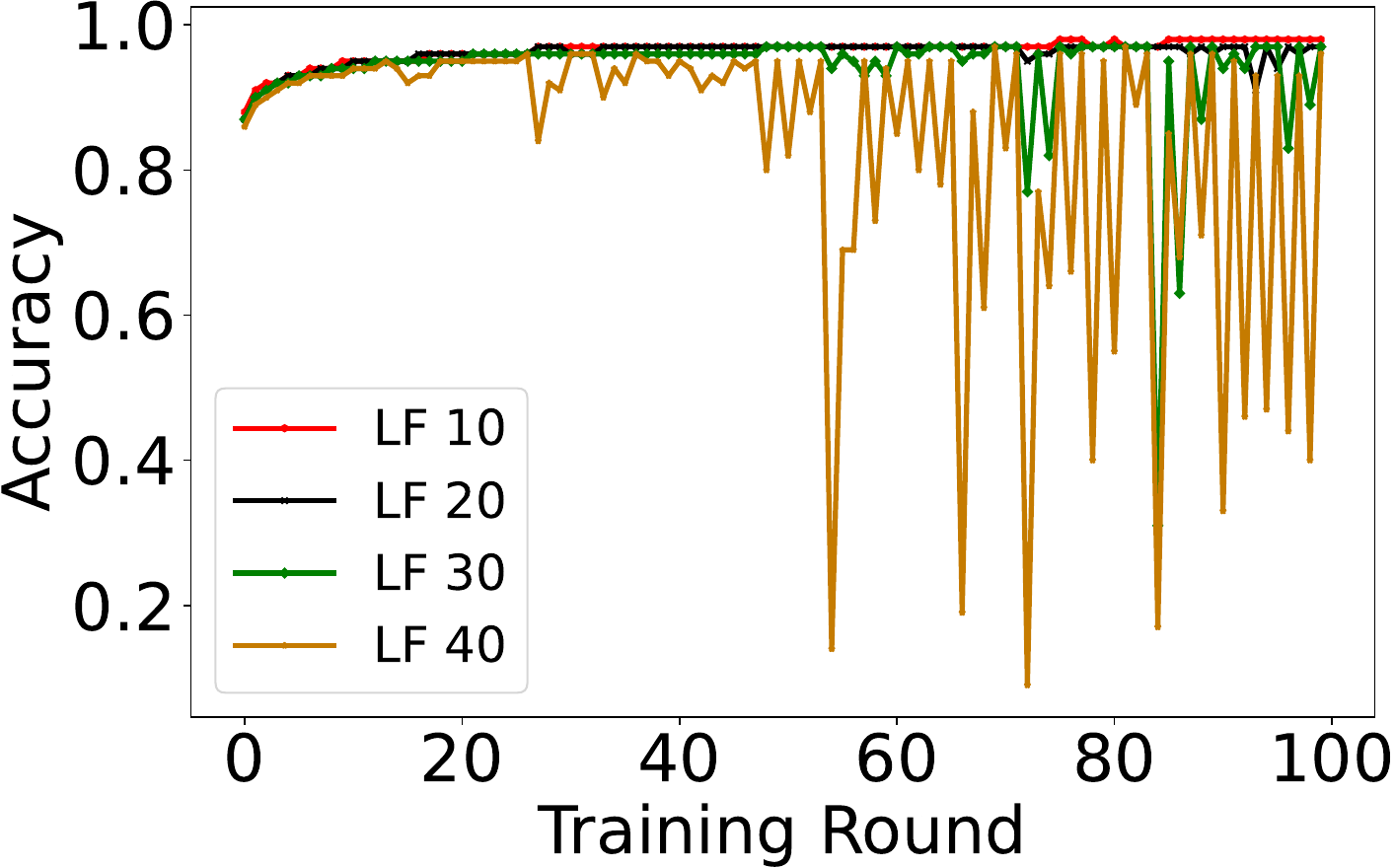}
    \caption*{(b) LF}
  \end{minipage}
  
  \caption{Evaluating the impact of increasing the number of PGA/LF attackers on \textbf{FedAvg}.}
  \label{fig:attacktest}
  \vspace{-2mm}
\end{figure}
\subsection{Analyzing attack model based on the number of attackers}
To find the number of attackers that significantly impact the performance of defense mechanisms across various scenarios, first, we examined the effects of different numbers of attackers on FedAvg in a simple scenario by using the MNIST dataset and a 2NN model in an IID setting, where each node has access to all classes in its training data. Fig. \ref{fig:attacktest} illustrates the impact of increasing the number of attackers on FedAvg over 100 training rounds. As expected, the PGA attack has a much stronger negative impact on the accuracy of the global model compared to the LF attack. For PGA, with 1 and 3 attackers (out of 100 nodes), FedAvg mitigates the attack's impact effectively, avoiding significant accuracy degradation. However, as shown in Fig. \ref{fig:attacktest}a, when the number of PGA attackers increases to 5, the accuracy of FedAvg fluctuates significantly, especially when an attacker node is selected in consecutive rounds for training. In contrast, for the LF attack (which is less severe than PGA), FedAvg demonstrates more resilience. For 10 and 20 LF attackers, FedAvg successfully manages the impact on the global model's accuracy. However, as the number of attackers increases to 30, the first noticeable drop in accuracy occurs, which further worsens as the number of attackers reaches 40. It should be noted that, although FedAvg was not initially designed as a defense mechanism, it exhibits substantial robustness against poisoning attacks \cite{sp22} and remains one of the most widely adopted AGR methods for federated learning. Based on these findings, our experiments focused on scenarios involving 5 and 10 PGA attackers, and 30 and 40 LF attackers.

\begin{table}[b]
\centering
\caption{The 2NN model architecture used in experiments for MNIST/FMNIST datasets.}
\begin{tabular}{|l|c|c|c|}
\hline
Layer & Output Shape & Activation & Parameters \\
\hline
Input & (28, 28, 1) & None & 0 \\
\hline
Dense & (200) & ReLU & 157,000 \\
\hline
Dense & (200) & ReLU & 40,200 \\
\hline
Dense & (10) & Softmax & 2,010 \\
\hline
\end{tabular}
\label{table:2nn}
\end{table}
\vspace{-2 mm}
\section{Evaluation Results}
To evaluate the performance of SHFL, we use TensorFlow library of Python to implement SHFL, Trimmed-mean \cite{trimmed-mean}, and Multi-krum \cite{multi-krum} defense mechanisms and FedAvg \cite{fedavg}, against PGA \cite{sp22} and LF \cite{lf} attacks in a network with 100 IoT nodes and 10 edge servers. We use FedAvg as the default AGR algorithm in our experiments  while Trimmed-mean and Multi-krum perform client selection for filtering out malicious models. We conducted our experiments on Amazon AWS EC2 $m6a.8xlarge$ instances by using MNIST, Fashion-MNIST (FMNIST), and CIFAR-10 datasets. We evaluate the impact of attacks on both CNN and 2NN models in IID and non-IID settings. For IID scenarios, we divide the training data into 100 shards with equal size and assign each shard to an IoT node. In non-IID scenarios, first, we sort the data by label, then partition the training data for each label into 10 equal shards and assign one shard to each IoT node. In both of these data distributions, each training data is assigned to only one node in the network. Tables \ref{table:2nn}, \ref{table:cnn1}, and \ref{table:cnn2} show the configurations of the neural network models we used in our experiments. All of these models are built by using $float64$ data type. For Multi-krum and Trimmed-mean, we use conventional 2 level hierarchy when all 100 clients are connected to the cloud server since these methods are designed for networks with one level model AGR. For SHFL, the model AGR is done at 2 levels when 10 IoT nodes are connected to an edge server. Table \ref{table:config} shows the configuration parameters of our experiments.
\vspace{-3 mm}
\subsection{Fidelity}
To examine the performance of SHFL in non-adversarial scenarios, first, we run experiments with different model architectures and learning rates to compare the performance of SHFL with vanilla FedAvg, which uses a single level of model AGR. Table \ref{table:fidelity-2nn} shows the accuracy of the global model for SHFL and FedAvg when the 2NN model is used with learning rates of 0.1 and 0.15. When learning rate is 0.1, SHFL consistently achieves higher accuracy by prioritizing client devices that make a more significant contribution to improve the accuracy of the global model. For experiments with learning rate 0.15, although SHFL’s accuracy is initially lower than FedAvg’s in the early rounds, by round 80, both methods achieve approximately the same results. Similar results are presented in Table \ref{table:fidelity-cnn} for the CNN model. At the 80th round, both methods achieve approximately the same results. After round 40, when FedAvg achieves higher accuracy compared to SHFL, the difference between the two methods is less than 0.5\%. As these results indicate, compared to traditional FL, performing AGR at two levels does not degrade the global model's accuracy (to two decimal points). These results demonstrate that SHFL can be effectively applied as a client selection mechanism in both adversarial and non-adversarial scenarios.

\begin{table}[t]
\caption{The CNN model architecture used in experiments for the MNIST dataset.}
\centering
\begin{tabular}{|l|c|c|c|}
\hline
Layer & Output Shape & Activation & Parameters \\
\hline
Input & (28, 28, 1) & None & 0 \\
\hline
Conv2D & (28, 28, 32) & None & 832 \\
\hline
MaxPooling2D & (14, 14, 32) & None & 0 \\
\hline
Conv2D & (14, 14, 64) & None & 51264 \\
\hline
MaxPooling2D & (7, 7, 64) & None & 0 \\
\hline
Flatten & (3136) & None & 0 \\
\hline
Dense & (512) & ReLU & 1606144 \\
\hline
Dense & (10) & Softmax & 5130 \\
\hline
\end{tabular}
\label{table:cnn1}
\end{table}

\begin{table}[t]
\caption{The CNN model architecture used in experiments for the CIFAR-10 dataset.}
\centering
\begin{tabular}{|l|c|c|c|}
\hline
Layer & Output Shape & Activation & Parameters\\
\hline
Input & (32, 32, 3) & None & 0 \\
\hline
Conv2D & (30, 30, 32) & ReLU & 896 \\
\hline
MaxPooling2D & (15, 15, 32) & None & 0 \\
\hline
Conv2D & (13, 13, 64) & ReLU & 18496 \\
\hline
MaxPooling2D & (6, 6, 64) & None & 0 \\
\hline
Conv2D & (4, 4, 64) & ReLU & 36928 \\
\hline
Flatten & (2304) & None & 0 \\
\hline
Dense & (64) & ReLU & 147520 \\
\hline
Dense & (10) & Softmax & 650 \\
\hline
\end{tabular}
\label{table:cnn2}
\end{table}

\begin{table}[t]
\caption{Configuration parameters of the experiments.}
\centering
\begin{tabular}{|l|c|c|c|}
\hline
Total number of client devices                 & 100\\
\hline
Number of edge servers                  & 10\\
\hline
Learning rate                   & 0.10 \\
\hline
Epoch                           & 5\\
\hline
Batch size                     & 32\\
\hline
Number of clients selected per training round                 & 30\\
\hline
Number of clients connected to an edge server                 & 10\\
\hline
Selection round for updating participating clients    & 3\\
\hline
\end{tabular}
\label{table:config}\vspace{-4 mm}
\end{table}

\begin{table}[b]
\caption{Accuracy of SHFL and FedAvg in a non-adversarial scenario for the 2NN model (MNIST dataset).}
\centering
\begin{tabular}{|l|c|c|c|c|}
\hline
\multirow{2}{*}{Round} & \multicolumn{2}{c|}{Learning Rate 0.1} & \multicolumn{2}{c|}{Learning Rate 0.15} \\
\cline{2-5}
 & FedAvg & SHFL & FedAvg & SHFL \\
\hline
20 & 0.7632 & 0.7749 & 0.7044 & 0.6767 \\
\hline
40 & 0.7837 & 0.7873 & 0.7705 & 0.8329 \\
\hline
60 & 0.8003 & 0.8589 & 0.8709 &  0.8663 \\
\hline
80 & 0.8501 & 0.8634 & 0.8706 & 0.8702 \\
\hline
\end{tabular}
\label{table:fidelity-2nn}
\end{table}

\begin{table}[b]
\caption{Accuracy of SHFL and FedAvg in a non-adversarial scenario for the CNN model (MNIST dataset).}
\centering
\begin{tabular}{|l|c|c|c|c|}
\hline
\multirow{2}{*}{Round} & \multicolumn{2}{c|}{Learning Rate 0.1} & \multicolumn{2}{c|}{Learning Rate 0.15} \\
\cline{2-5}
 & FedAvg & SHFL & FedAvg & SHFL \\
\hline
20 & 0.8207 & 0.7794 & 0.8539 &  0.7351 \\
\hline
40 & 0.8965 & 0.8296 & 0.9242  &  0.9148  \\
\hline
60 & 0.9269 & 0.9271 & 0.9388 &  0.9357 \\
\hline
80 & 0.9326 & 0.9391 & 0.9506 & 0.9496 \\
\hline
\end{tabular}
\label{table:fidelity-cnn}
\end{table}

\vspace{-2 mm}
\subsection{PGA Attack}

\begin{figure*}[!ht]
  \centering
  \subfloat[ \normalfont{IID (5 attackers)}]{%
      \includegraphics[width= 40 mm, height=35 mm]{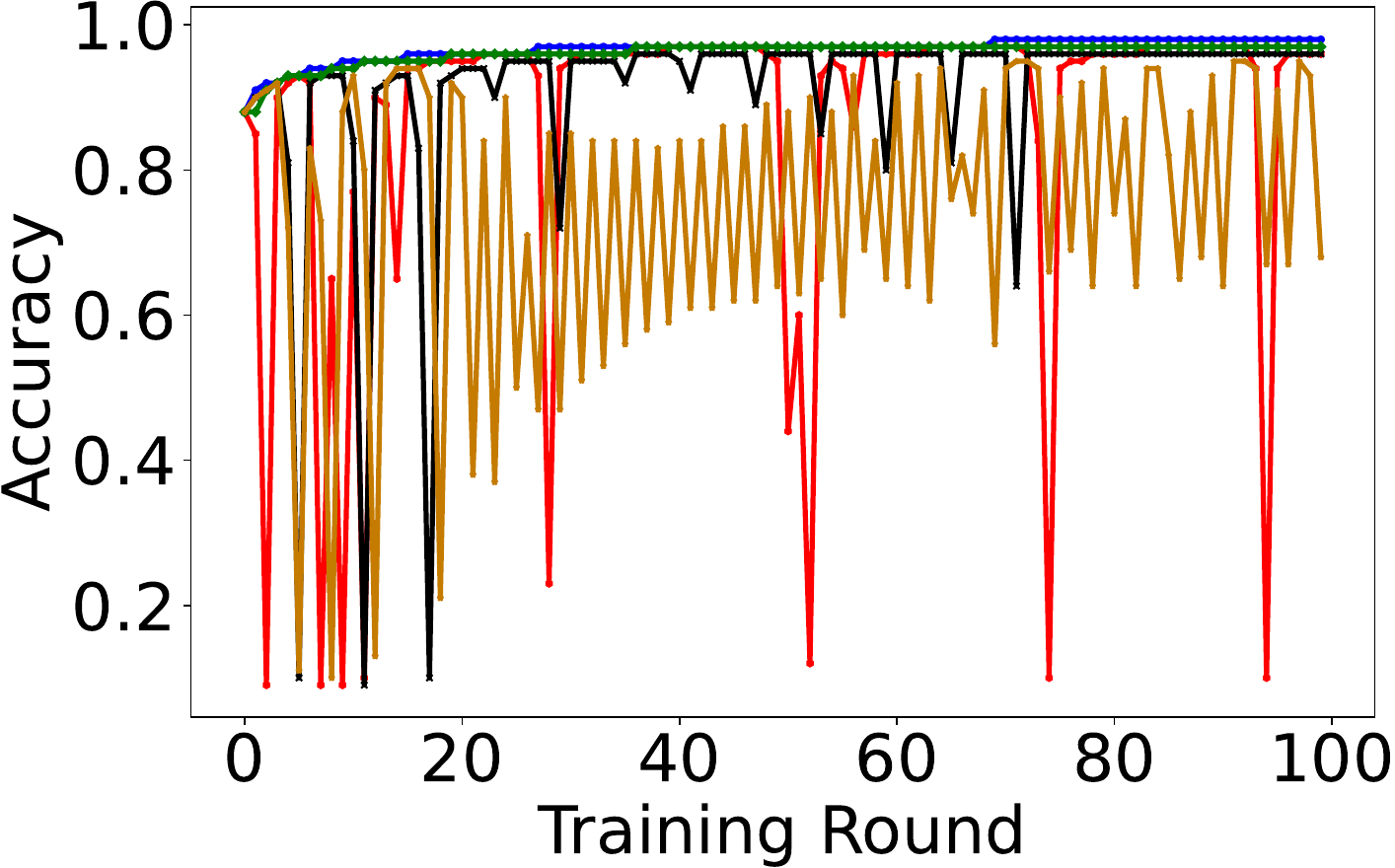}%
      \label{fig:mnist-2nn-iid-pga5}}\hspace{2 mm}
   \subfloat[\normalfont{IID (10 attackers)}]{%
      \includegraphics[width= 40 mm, height=35 mm]{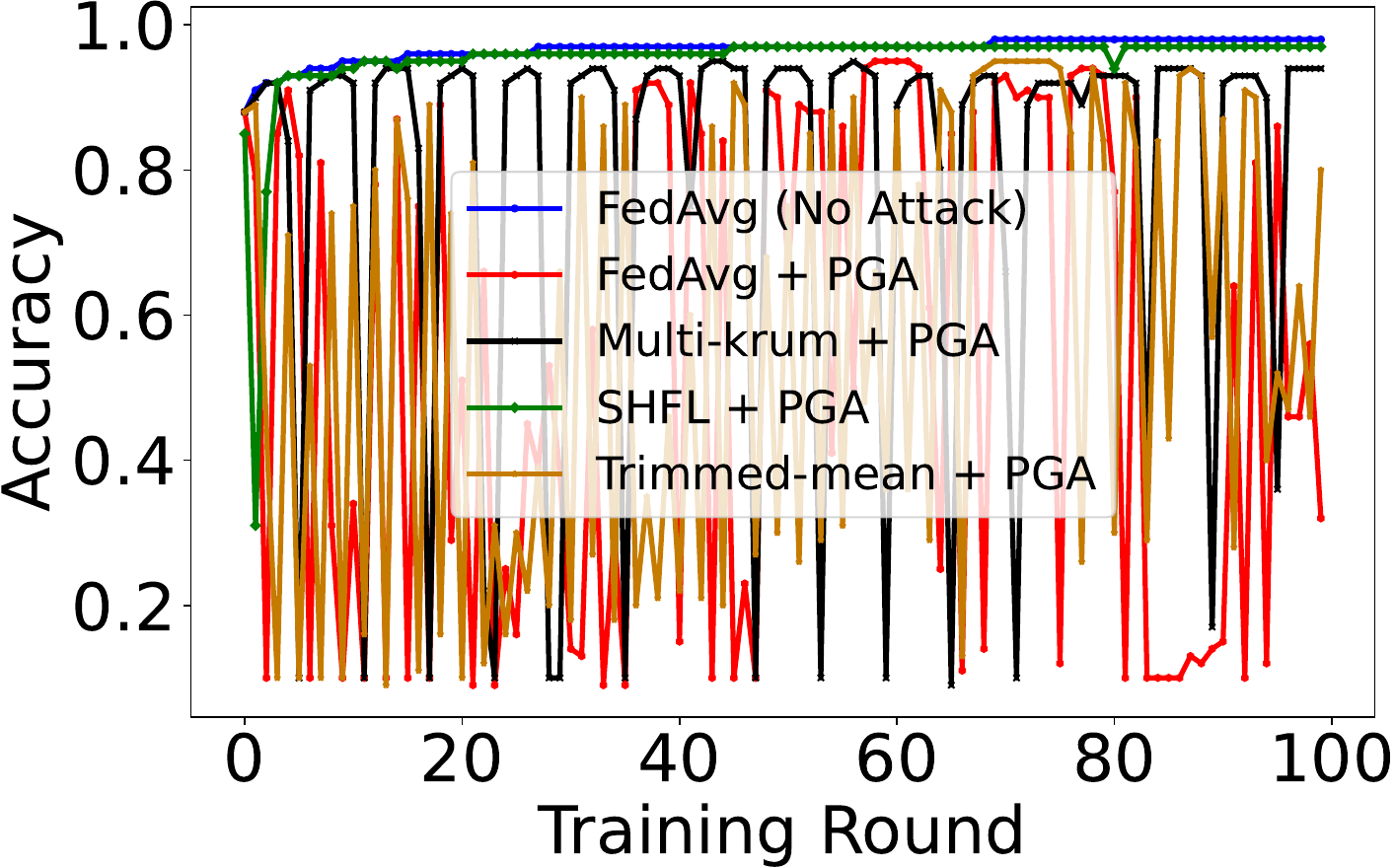}%
      \label{fig:mnist-2nn-iid-pga10}}\hspace{2 mm}
   \subfloat[\normalfont{non-IID (5 attackers)}]{%
      \includegraphics[width= 40 mm, height=35 mm]{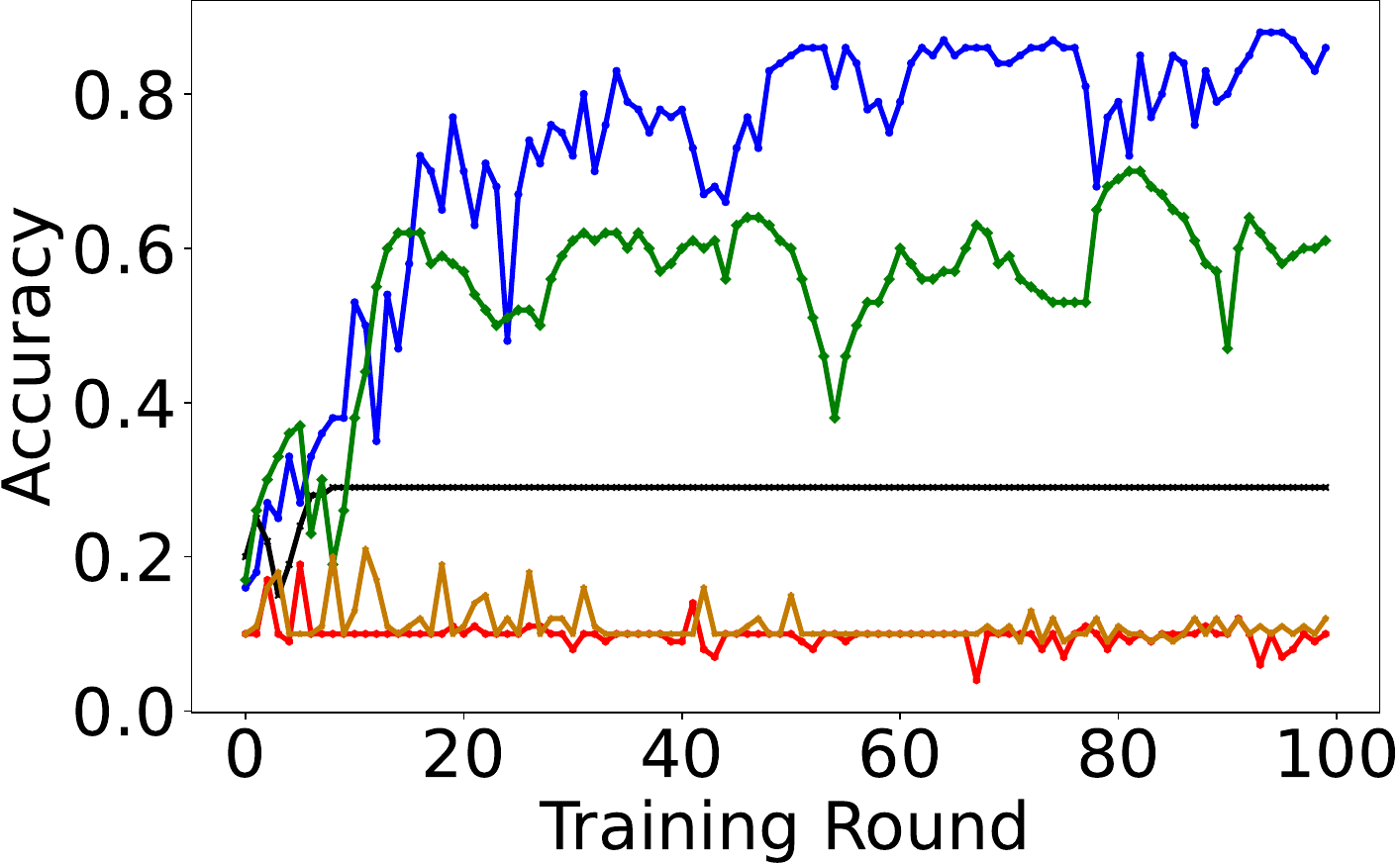}%
      \label{fig:mnist-2nn-noniid-pga5}}\hspace{2 mm}
   \subfloat[\normalfont{non-IID (10 attackers)}]{%
      \includegraphics[width= 40 mm, height=35 mm]{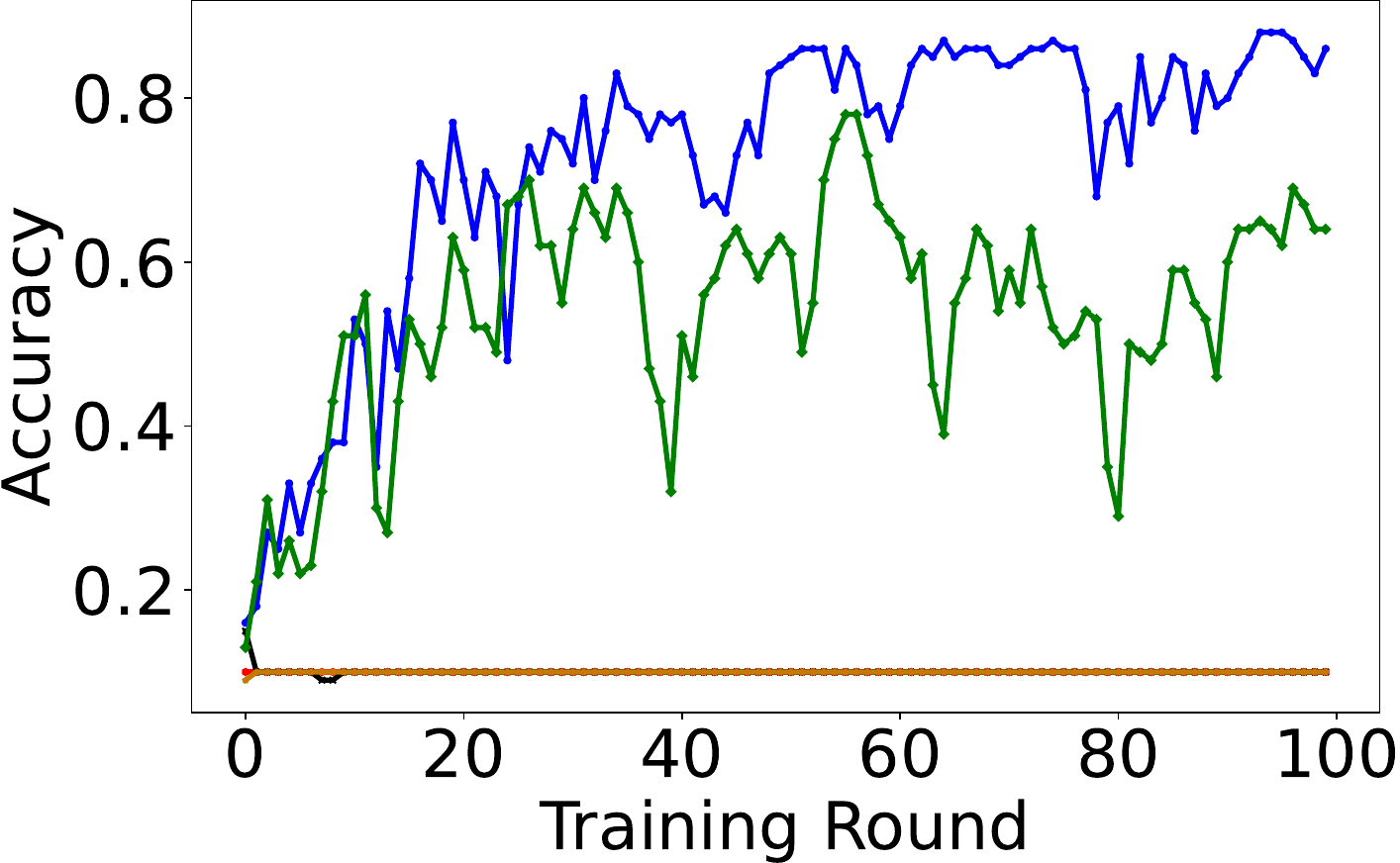}%
      \label{fig:mnist-2nn-noniid-pga10}}

  \caption{Accuracy of defense mechanisms using the \textbf{2NN} model on \textbf{MNIST} dataset for IID and non-IID scenarios under \textbf{PGA attack} (5 and 10 attacker nodes).}
  \label{fig:mnist-2nn-pga}\vspace{-2 mm}
\end{figure*}

\begin{figure*}[h!]
  \centering
  \subfloat[ \normalfont{IID (5 attackers)}]{%
      \includegraphics[width= 40 mm, height=35 mm]{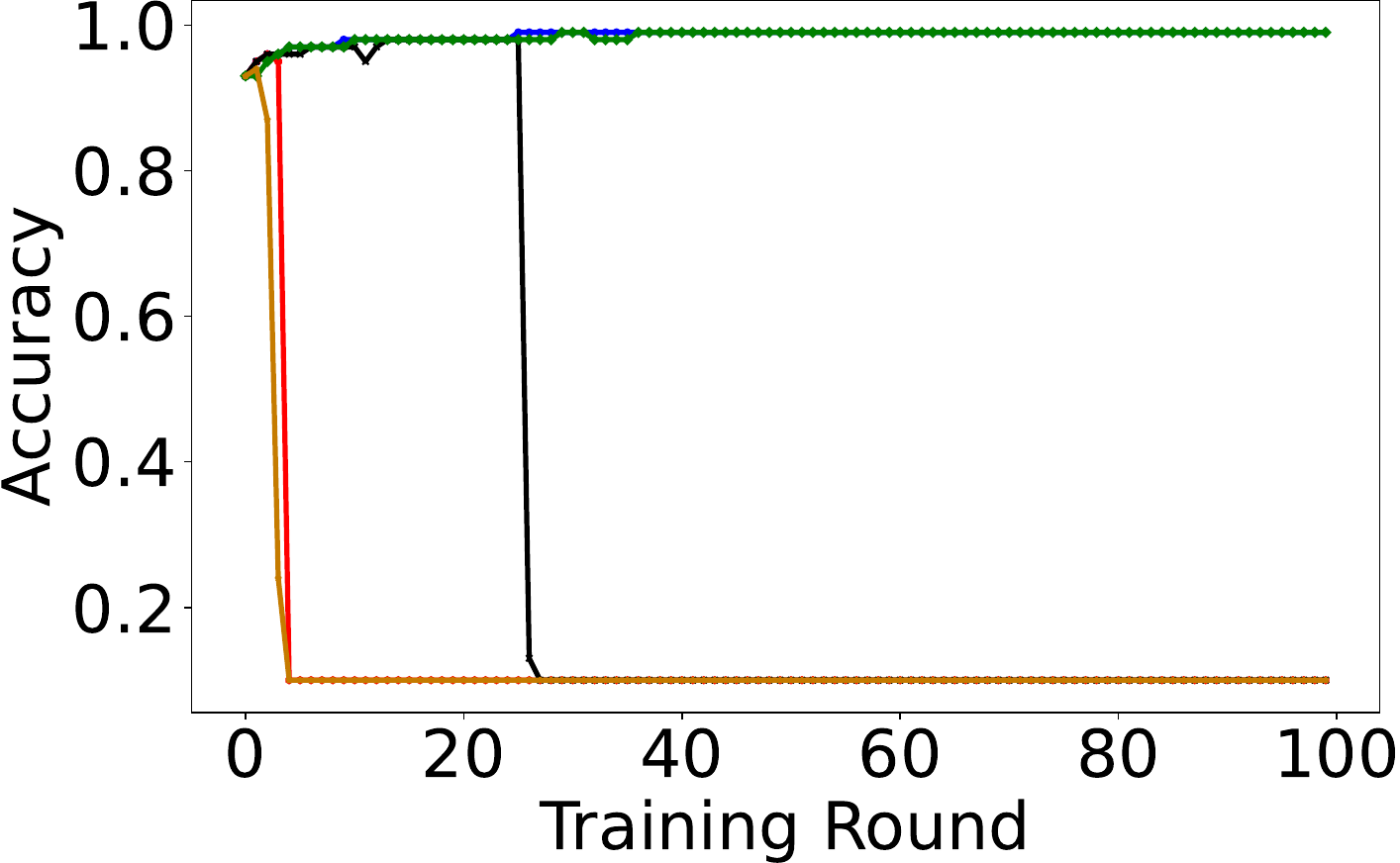}%
      \label{fig:mnist-cnn-iid-pga5}}\hspace{2 mm}
   \subfloat[\normalfont{IID (10 attackers)}]{%
      \includegraphics[width= 40 mm, height=35 mm]{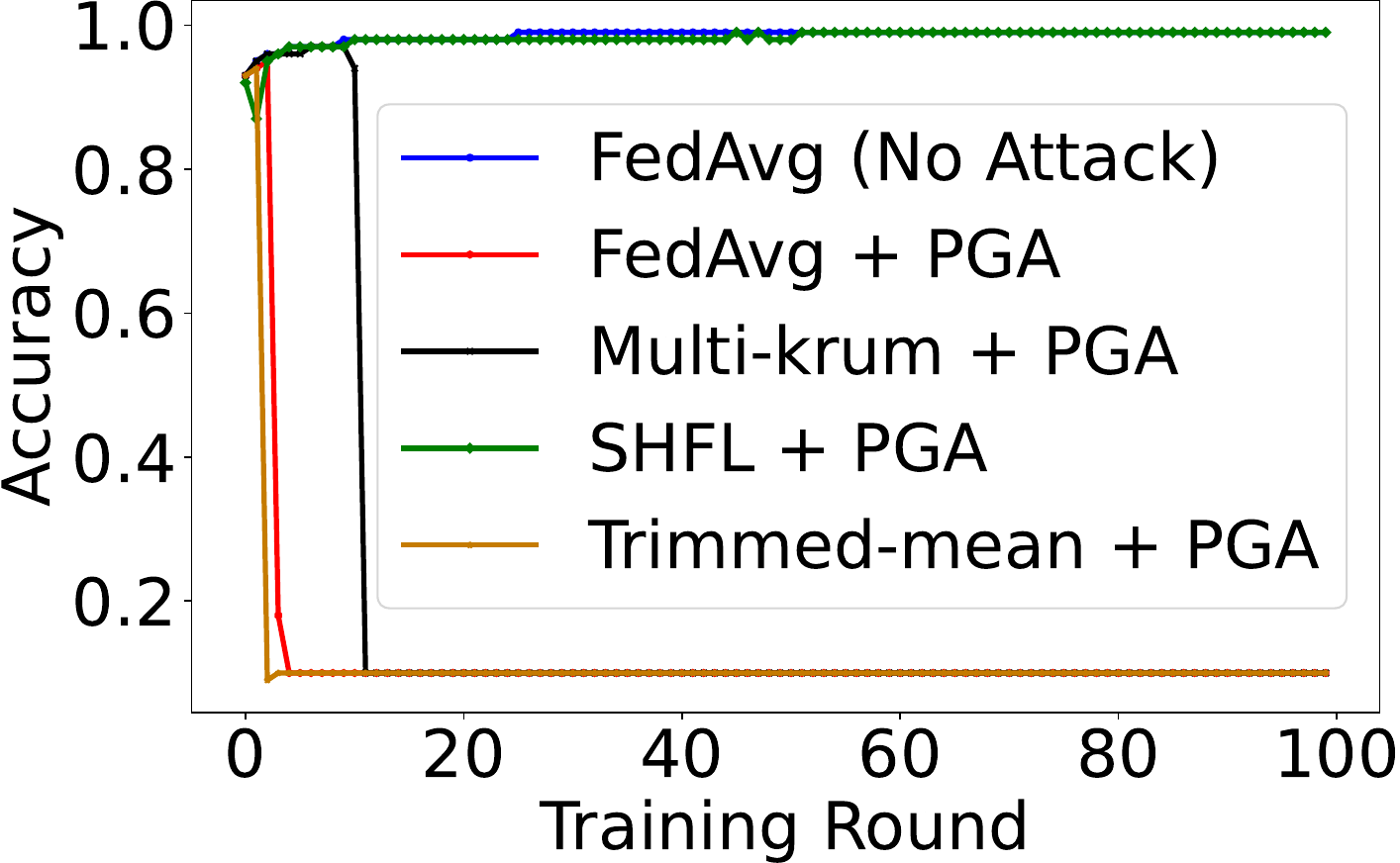}%
      \label{fig:mnist-cnn-iid-pga10}}\hspace{2 mm}
   \subfloat[\normalfont{non-IID (5 attackers)}]{%
      \includegraphics[width= 40 mm, height=35 mm]{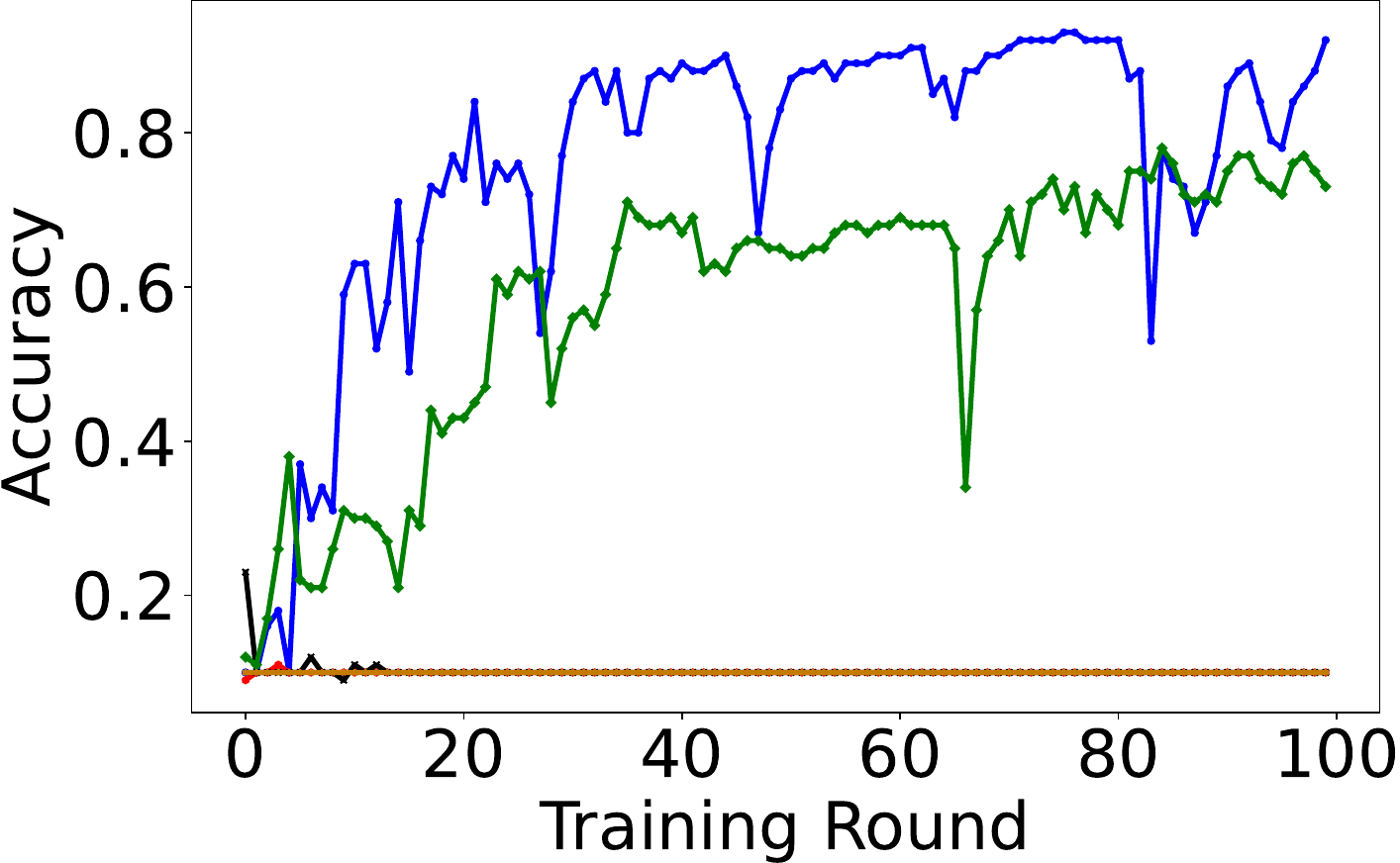}%
      \label{fig:mnist-cnn-noniid-pga5}}\hspace{2 mm}
   \subfloat[\normalfont{non-IID (10 attackers)}]{%
      \includegraphics[width= 40 mm, height=35 mm]{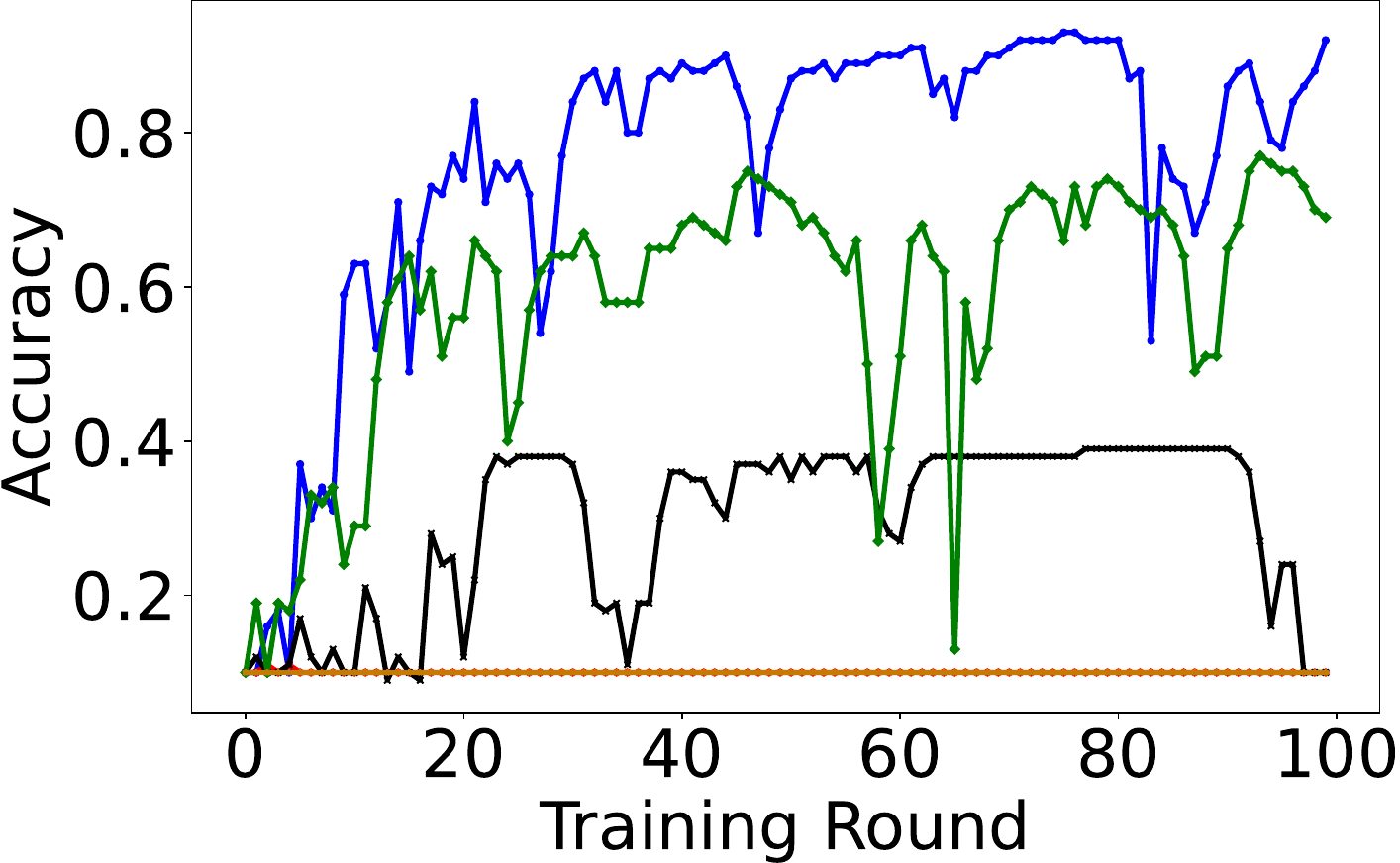}%
      \label{fig:mnist-cnn-noniid-pga10}}

  \caption{Accuracy of defense mechanisms using the \textbf{CNN} model on \textbf{MNIST} dataset for IID and non-IID scenarios under \textbf{PGA attack} (5 and 10 attacker nodes).}
  \label{fig:mnist-cnn-pga}\vspace{-2 mm}
\end{figure*}

\begin{figure*}[h!]
  \centering
  \subfloat[ \normalfont{IID (5 attackers)}]{%
      \includegraphics[width= 40 mm, height=35 mm]{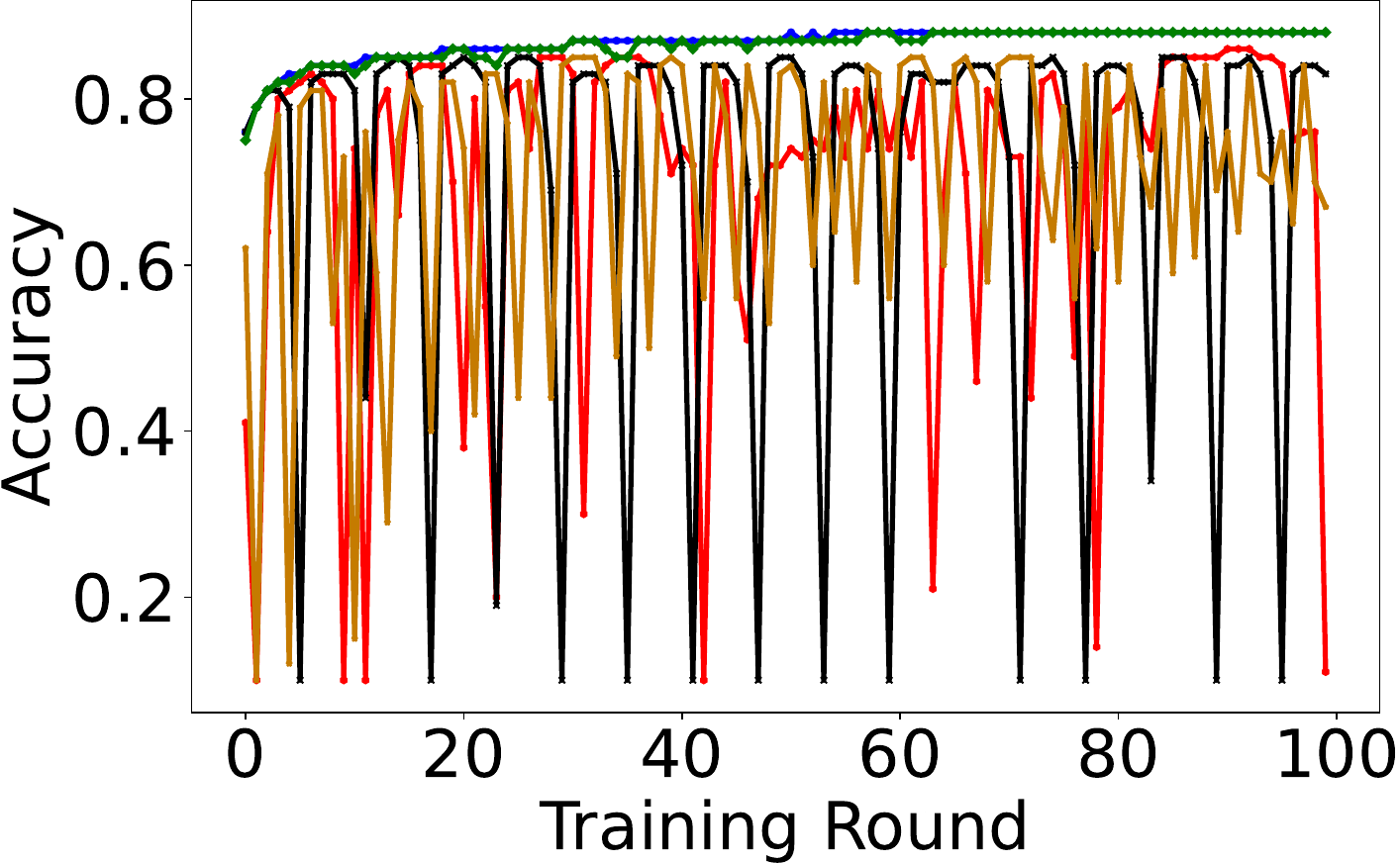}%
      \label{fig:fmnist-iid-pga5}}\hspace{2 mm}
   \subfloat[\normalfont{IID (10 attackers)}]{%
      \includegraphics[width= 40 mm, height=35 mm]{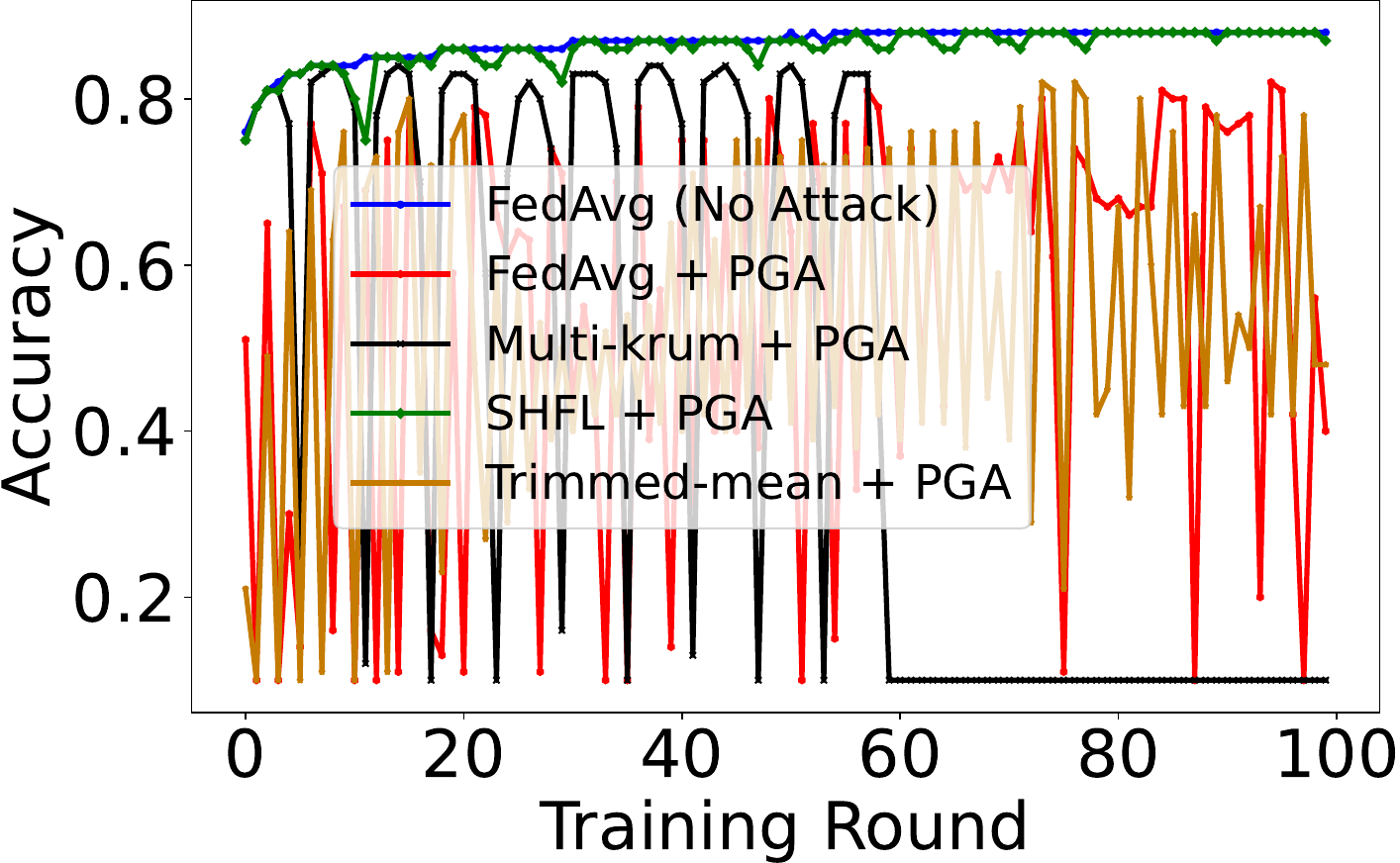}%
      \label{fig:fmnist-iid-pga10}}\hspace{2 mm}
   \subfloat[\normalfont{non-IID (5 attackers)}]{%
      \includegraphics[width= 40 mm, height=35 mm]{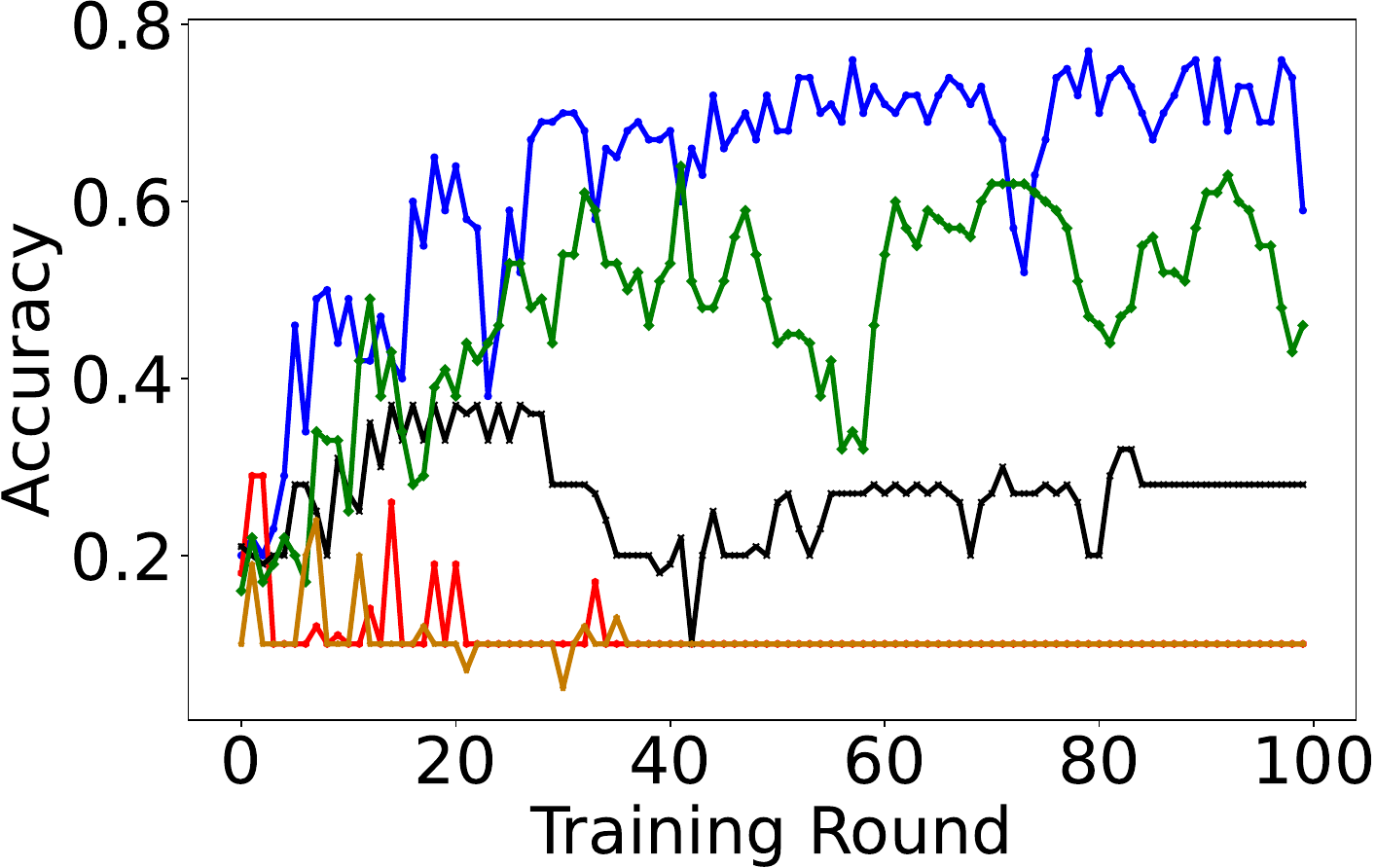}%
      \label{fig:fmnist-noniid-pga5}}\hspace{2 mm}
   \subfloat[\normalfont{non-IID (10 attackers)}]{%
      \includegraphics[width= 40 mm, height=35 mm]{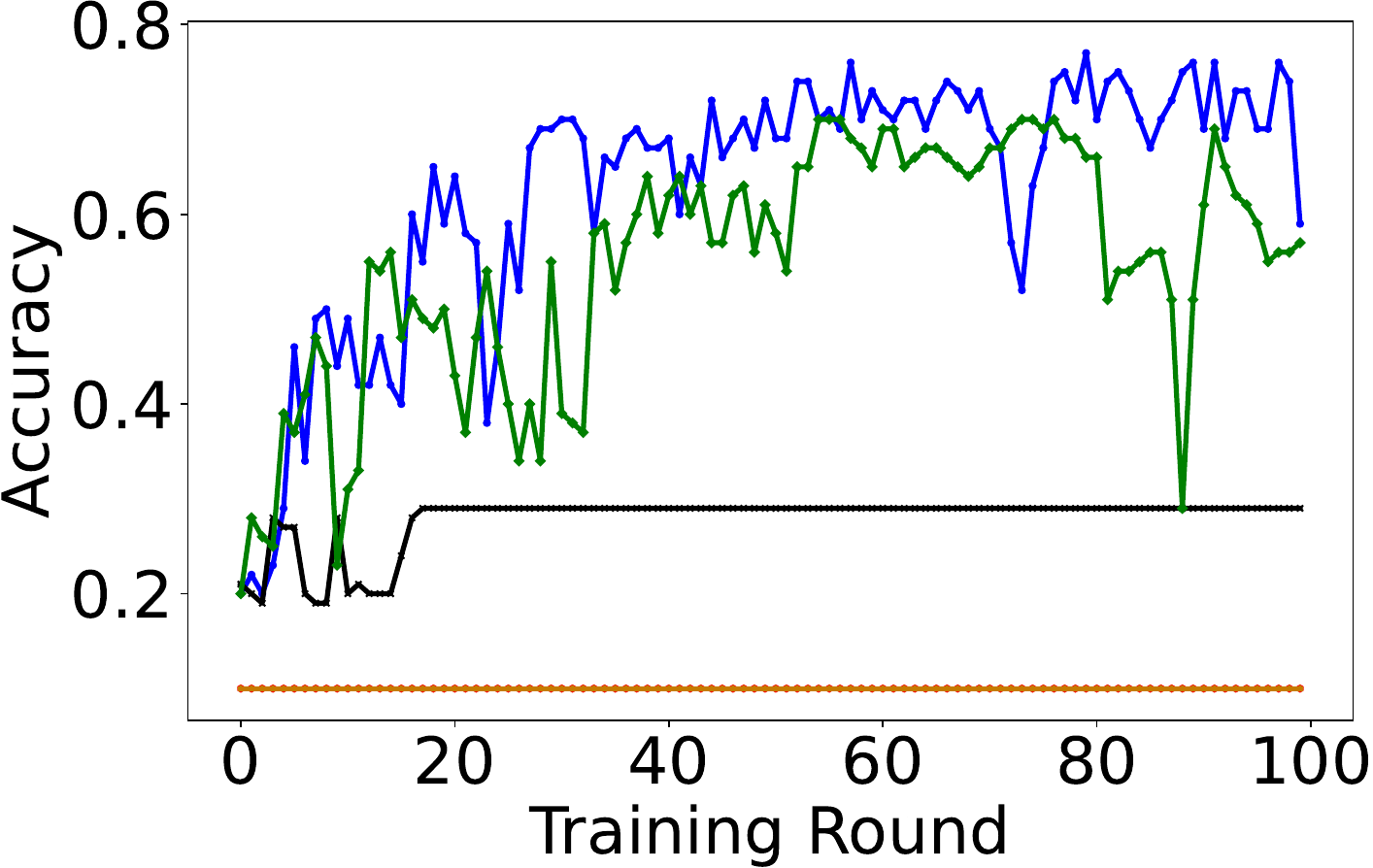}%
      \label{fig:fmnist-noniid-pga10}}

  \caption{Accuracy of defense mechanisms using the \textbf{2NN} model on \textbf{FMNIST} dataset for IID and non-IID scenarios under \textbf{PGA attack} (5 and 10 attacker nodes).}
  \label{fig:fmnist-pga}\vspace{-2 mm}
\end{figure*}

\begin{figure*}[h!]
  \centering
  \subfloat[ \normalfont{IID (5 attackers)}]{%
      \includegraphics[width= 40 mm, height=35 mm]{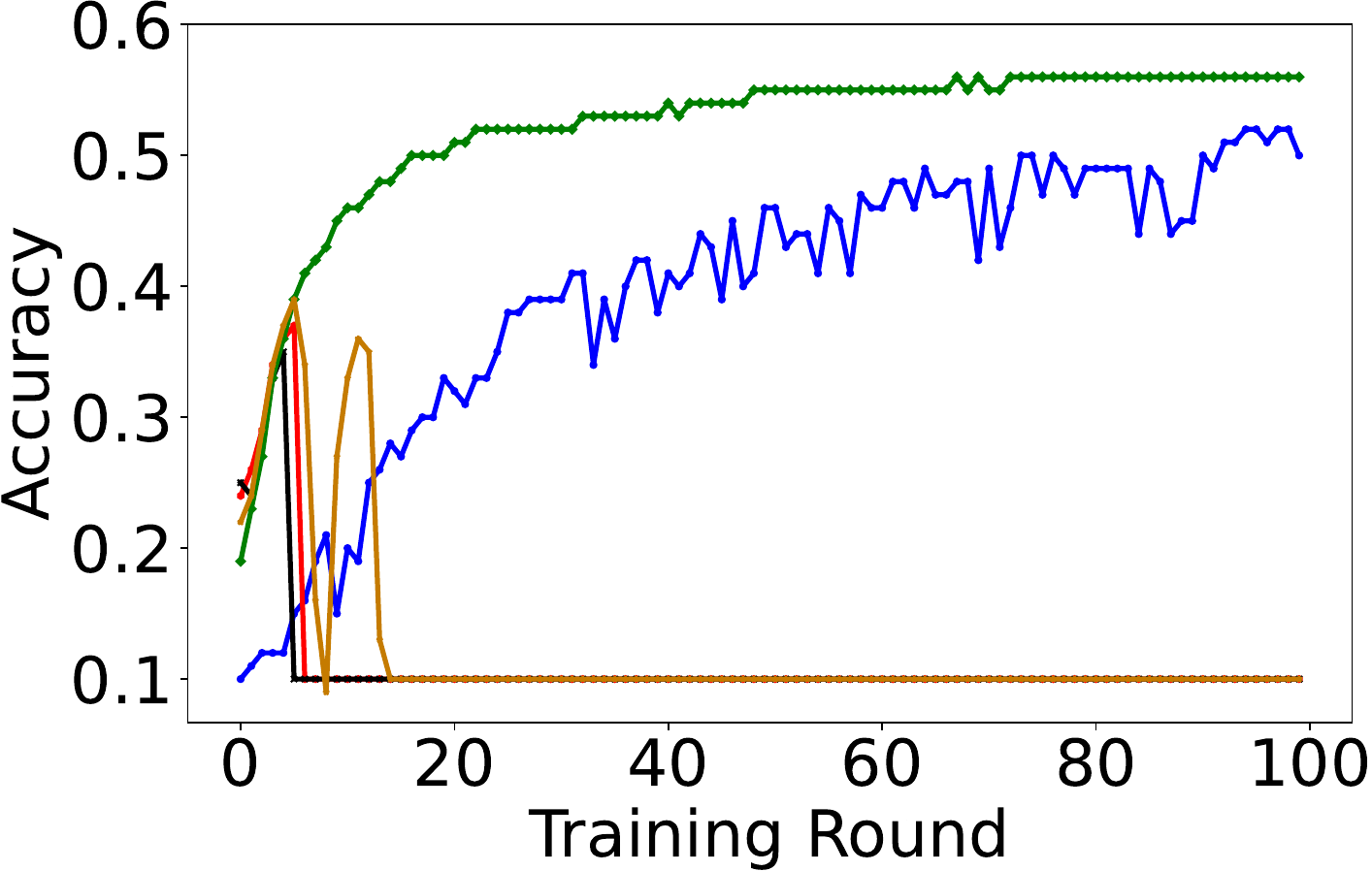}%
      \label{fig:cifar-iid-pga5}}\hspace{2 mm}
   \subfloat[\normalfont{IID (10 attackers)}]{%
      \includegraphics[width= 40 mm, height=35 mm]{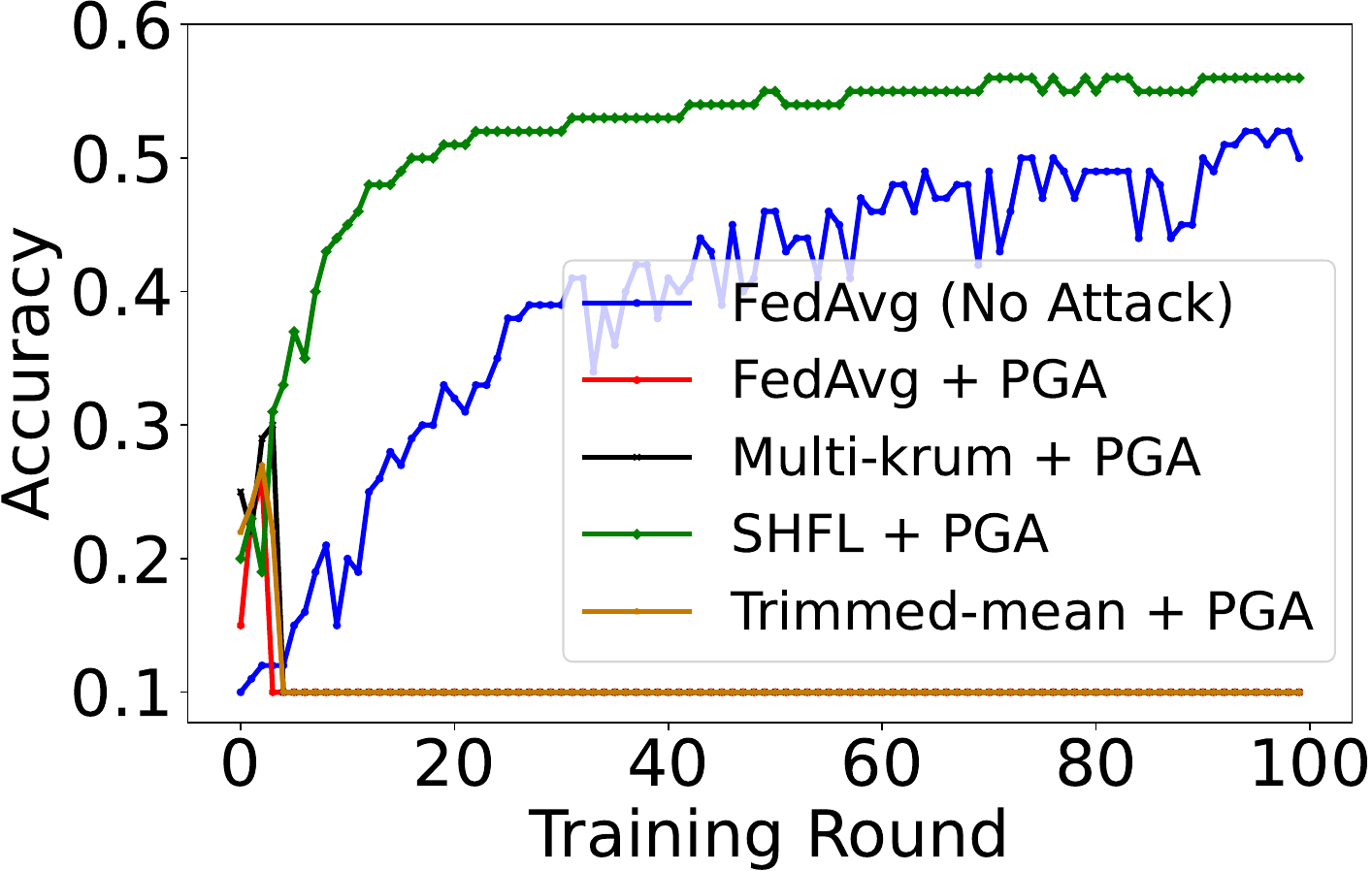}%
      \label{fig:cifar-iid-pga10}}\hspace{2 mm}
   \subfloat[\normalfont{non-IID (5 attackers)}]{%
      \includegraphics[width= 40 mm, height=35 mm]{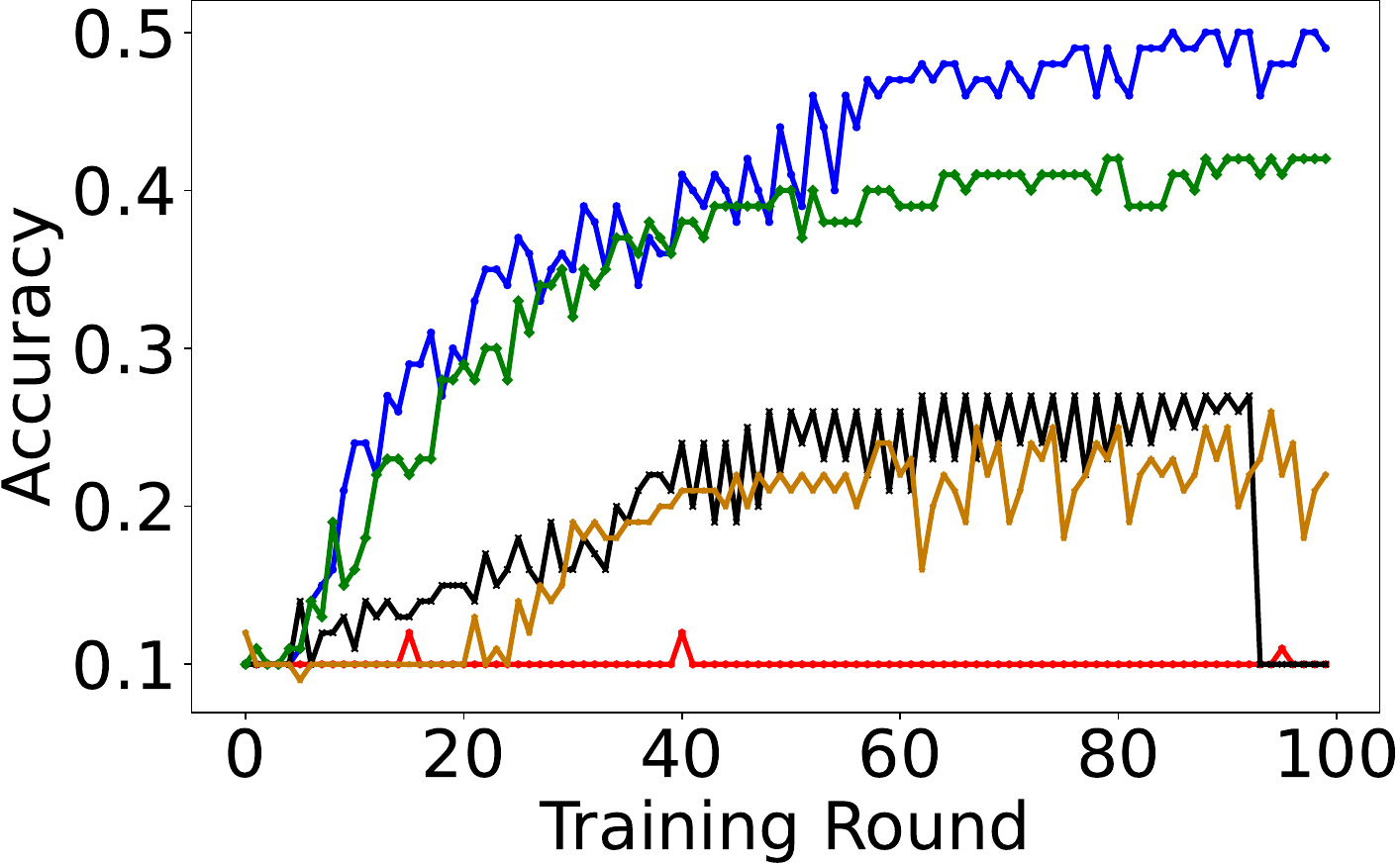}%
      \label{fig:cifar-noniid-pga5}}\hspace{2 mm}
   \subfloat[\normalfont{non-IID (10 attackers)}]{%
      \includegraphics[width= 40 mm, height=35 mm]{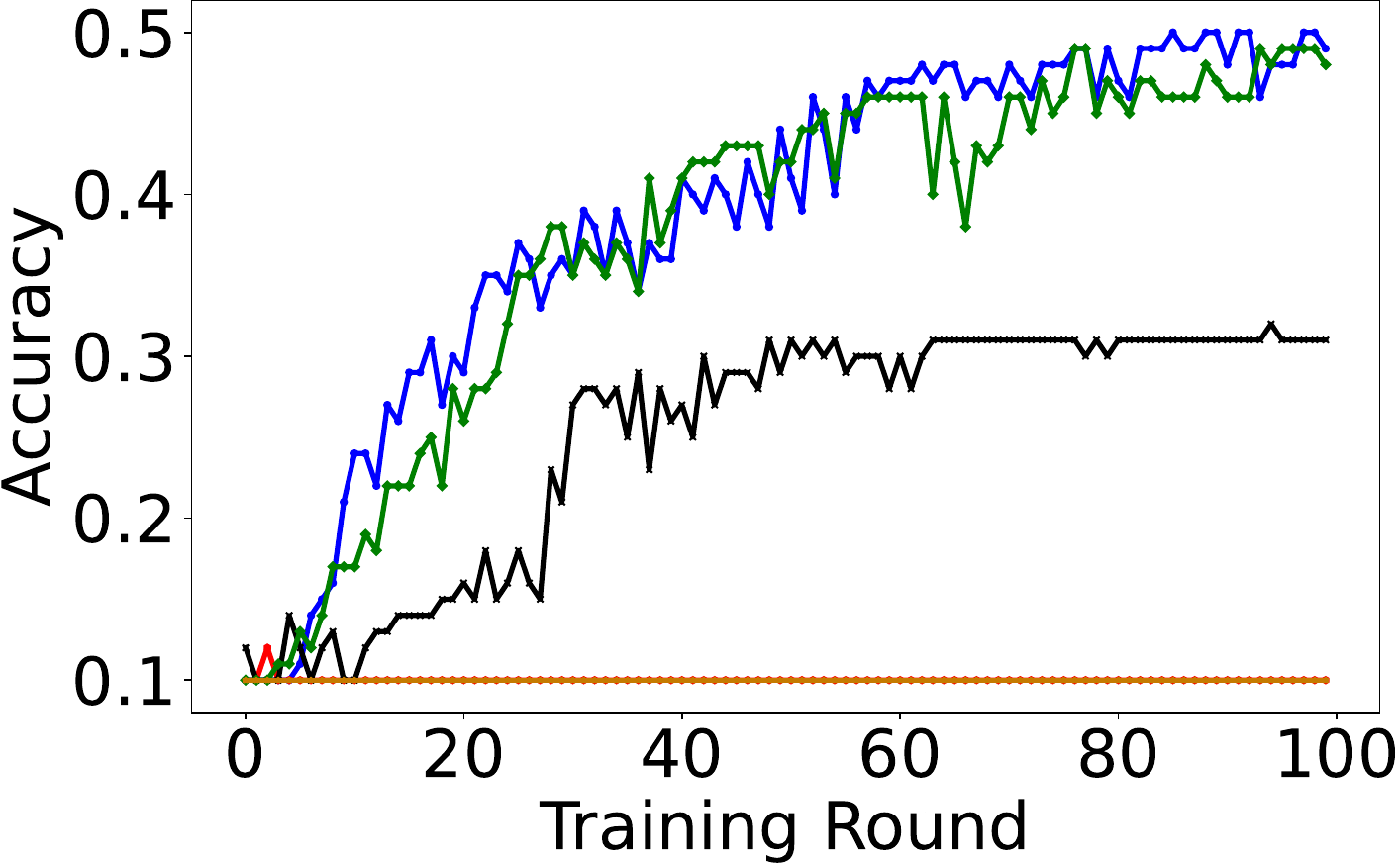}%
      \label{fig:cifar-noniid-pga10}}

  \caption{Accuracy of defense mechanisms using the \textbf{CNN} model on \textbf{CIFAR-10} dataset for IID and non-IID scenarios under \textbf{PGA attack} (5 and 10 attacker nodes).}
  \label{fig:cifar-pga}\vspace{-4 mm}
\end{figure*}

In the first set of experiments, we assess the performance of SHFL in the presence of 5 or 10 PGA attackers (Figs. \ref{fig:mnist-2nn-pga}, \ref{fig:mnist-cnn-pga}, \ref{fig:fmnist-pga}, and \ref{fig:cifar-pga}), randomly selected from all client devices. Fig. \ref{fig:mnist-2nn-pga} shows the accuracy of defense mechanisms for both IID and non-IID scenarios by using the MNIST dataset and the 2NN model. In the IID setting (Figs. \ref{fig:mnist-2nn-iid-pga5} and \ref{fig:mnist-2nn-iid-pga10}), all defense mechanisms, along with FedAvg, achieve maximum accuracies (Table \ref{table:trace}) higher than 94\%. However, except SHFL, other defense methods experience noticeable drops in accuracy whenever an attacker node is selected in the client selection process. This instability is especially pronounced in Figure \ref{fig:mnist-2nn-iid-pga10}, where the presence of 10 PGA attackers causes greater performance degradation. In contrast, SHFL demonstrates remarkable stability. Its accuracy never drops below 94\%, thanks to its ability to mitigate the impact of attacker-connected edge models in the AGR process running at the cloud for creating the global model. Moreover, among all defense mechanisms, SHFL achieves the highest accuracy (97)\%. In the non-IID setting, the impact of PGA attacks on the global model is significantly more severe. As shown in Figs. \ref{fig:mnist-2nn-noniid-pga5} and \ref{fig:mnist-2nn-noniid-pga10}, the maximum accuracy of the other defense mechanisms does not exceed 30\%, while SHFL achieves accuracy over 70\%. Our experiments reveal that once Multi-krum or Trimmed-mean selects an attacker node during client selection, the PGA attack corrupts the global models by dramatically increasing its gradients, preventing recovery to a stable state. 

\begin{table*}[t]
\centering
\caption{The maximum accuracy of different defense mechanisms for IID and non-IID scenarios in 100 training rounds in the presence of 5 and 10 PGA attackers.}\vspace{-2 mm}
{ 
\begin{tabular}{|c|c|c|c|c|c|c|c|c|c|c|}
\hline
\multirow{3}{*}{Defense} & \multicolumn{4}{c|}{MNIST} & \multicolumn{2}{c|}{FMNIST} & \multicolumn{2}{c|}{CIFAR-10} \\ \cline{2-9} 
                                  & \multicolumn{2}{c|}{CNN}   & \multicolumn{2}{c|}{2NN}   & \multicolumn{2}{c|}{2NN}  & \multicolumn{2}{c|}{CNN}   \\ \cline{2-9} 
                                  & IID      & non-IID   & IID      & non-IID   & IID    & non-IID & IID     & non-IID \\ \hline
FedAvg (No Attack)                & 0.99              & 0.92              & 0.97              & 0.88              & 0.88            & 0.76            & 0.52             & 0.50             \\ \hline
FedAvg (5)                  & 0.95              & 0.11              & 0.96              & 0.18              & 0.85            & 0.29            & 0.37             & 0.12             \\ \hline
FedAvg (10)                 & 0.95              & 0.11              & 0.96              & 0.10              & 0.81            & 0.10            & 0.25             & 0.11             \\ \hline
Multi-krum (5)                 & 0.97              & 0.23              & 0.96              & 0.29              & 0.84            & 0.37            & 0.35             & 0.27             \\ \hline
Multi-krum (10)                & 0.97              & 0.38              & 0.96              & 0.15              & 0.84            & 0.29            & 0.29             & 0.31             \\ \hline
SHFL (5)                 & 0.98              & 0.77              & 0.97              & 0.70              & 0.88            & 0.63            & 0.56             & 0.42             \\ \hline
SHFL (10)                & 0.98              & 0.76              & 0.97              & 0.78              & 0.88            & 0.70            & 0.56             & 0.49             \\ \hline
Trimmed-mean (5)         & 0.93              & 0.10              & 0.95              & 0.20              & 0.85            & 0.24            & 0.39             & 0.25             \\ \hline
Trimmed-mean (10)        & 0.94              & 0.10              & 0.95              & 0.10              & 0.82            & 0.10            & 0.26             & 0.10             \\ \hline
\end{tabular}\label{table:trace}
} 
\end{table*}

\begin{table}[b]
\centering
\vspace{-3 mm}\caption{The weight of each edge model in the SHFL's AGR process ($\zeta=0.1 , \tau=1$) running at the cloud when 2 attackers are selected at edge networks 3 and 4 for training.}
\begin{tabular}{|c|c|c|c|}
\hline
Edge network & Round 0 & Round 1 & Round 2 \\ \hline
Edge 1 & 0.8237 & 0.8958 & 1.2084 \\ \hline
Edge 2 & 0.8151 & 0.8099 & 1.0999 \\ \hline
Edge 3 & 1.7023 & 1.4191 & \textbf{0.1} \\ \hline
Edge 4 & 1.7163 & 1.5082 & \textbf{0.1} \\ \hline
Edge 5 & 0.8156 & 0.8433 & 1.2091 \\ \hline
Edge 6 & 0.8457 & 0.9173 & 1.2450 \\ \hline
Edge 7 & 0.8151 & 0.9044 & 1.2394 \\ \hline
Edge 8 & 0.8232 & 0.8437 & 1.1351 \\ \hline
Edge 9 & 0.8301 & 1.0151 & 1.5261 \\ \hline
Edge 10 & 0.8129 & 0.8433 & 1.1369 \\ \hline
\end{tabular}\label{table:shfl}\vspace{-3 mm}
\end{table}

In the experiments with the CNN model (Fig. \ref{fig:mnist-cnn-pga}), we observed that Multi-krum fails to recover the global model when it consecutively selects PGA attackers across multiple rounds. In this case, the global model’s loss values became increasingly unstable. As shown in Fig. \ref{fig:mnist-cnn-iid-pga5}, at training rounds 26, the accuracy of Multi-krum suddenly drops from 97.61\% to 14.14\%. This is because the PGA attackers manipulate the gradients by significantly altering their values. When an attacker updates the edge server with a poisoned model that has extreme gradients, over several rounds, invalid mathematical operations occur in the AGR process for values exceeding numerical limits. In both Figs. \ref{fig:mnist-cnn-iid-pga5} and \ref{fig:mnist-cnn-iid-pga10}, although Multi-krum and Trimmed-mean achieve accuracy higher than 90\% in the initial rounds, their accuracy drops to around 10\% in less than 30 rounds. On the other hand, by filtering attacker nodes at two AGR stages (edge and cloud), SHFL first avoids selecting attacker nodes at the edge level and then reduces the impact of the edge models connected to attackers at the cloud level. As a result, it maintains its accuracy higher than 80\% in all training rounds. Similar to the results of the 2NN model, in non-IID distributions (Figs. \ref{fig:mnist-cnn-noniid-pga5} and \ref{fig:mnist-cnn-noniid-pga10}), the strength of the PGA attacks is amplified. For 10 PGA attackers, although SHFL experiences drops in accuracy, it manages to recover and achieve performance above 60\%. To demonstrate how SHFL prevents poisoned updates from corrupting the global model at the cloud, we present a part of our trace log for SHFL in Table \ref{table:shfl}, which shows the weight of each edge model in the AGR process at the cloud in three training rounds where two attackers connected to edge servers 3 and 4, are among participant. With $\zeta=0.1$ and $\tau=10$ (SHFL configuration), at round 2, the weights of edge models 3 and 4 are reduced to $0.1$ (the minimum value) in the AGR process. This enables SHFL to recover even after selecting an attacker node at the edge for training.

Fig. \ref{fig:fmnist-pga} illustrates the performance of various defense mechanisms against the PGA attack on the FMNIST dataset. In the IID setting, similar to experiments for the MNIST dataset, FedAvg, Multi-krum, and Trimmed-mean suffer significant accuracy drops when a participating client is compromised by the PGA attack. However, due to the simplicity of the dataset, these methods can partially mitigate the attack’s impact on the global model after several training rounds. When the number of attackers increases to 10, after 60 rounds, the accuracy of Multi-krum drops to 10\% and fails to recover, indicating that PGA has successfully corrupted the global model by pushing its gradients beyond acceptable limits. In the non-IID scenario, Trimmed-mean exhibits the worst performance, particularly during the early training rounds, as it struggles to detect attackers. This is due to PGA’s norm-bounding mechanism, which constrains the attacker's model update to have an $L_{2}$-norm similar to that of the global model, making it difficult for Trimmed-mean to filter out the malicious updates. Since Trimmed-mean relies on outlier detection, the norm-bounded updates from the attackers bypass its defenses. As shown in Table \ref{table:trace},  neither Multi-krum, FedAvg, nor Trimmed-mean manage to achieve 30\% accuracy in 100 rounds. In contrast, SHFL reaches a maximum accuracy of 70\%, thanks to its AGR process at the cloud level, which effectively detects poisoned edge models and minimizes their impacts on the process of creating a new global model.

For CIFAR-10 dataset (Fig. \ref{fig:cifar-pga}), the increased complexity of the data and the larger model size significantly amplify the performance gap between SHFL and other defense mechanisms. In the IID setting, within fewer than 20 rounds, FedAvg, Multi-krum, and Trimmed-mean fail to withstand the PGA attack, which progressively corrupts the global model, driving its accuracy down to around 10\% and rendering it unusable. In the non-IID scenario, although Multi-krum attempts to exclude PGA attackers through its client selection process, it struggles to improve accuracy beyond 32\%. This is due to the fairness problem inherent in Multi-krum, which becomes particularly evident when only a subset of clients (e.g., 30\%) are selected for training. Multi-krum prioritizes updates that are closest to the majority in terms of Euclidean distance, leading to a biased selection process where certain benign clients with models closer to the global model are consistently chosen, while benign clients with slightly different updates are excluded. In other words, while Multi-krum avoids selecting malicious updates, it also neglects important data from benign clients whose updates deviate from the majority. This limited diversity in the training data keeps the accuracy stagnant across rounds. In contrast, SHFL addresses this issue by first filtering out potentially malicious models and then randomly selecting among the remaining benign clients, ensuring a more balanced and diverse update process. Table \ref{table:trace} shows the maximum accuracy that each of defense methods achieves against the PGA attack in 100 training rounds. \vspace{-3 mm}

\subsection{LF Attack}

\begin{figure*}[h!]
  \centering
  \subfloat[ \normalfont{IID (30 attackers)}]{%
      \includegraphics[width= 40 mm, height=35 mm]{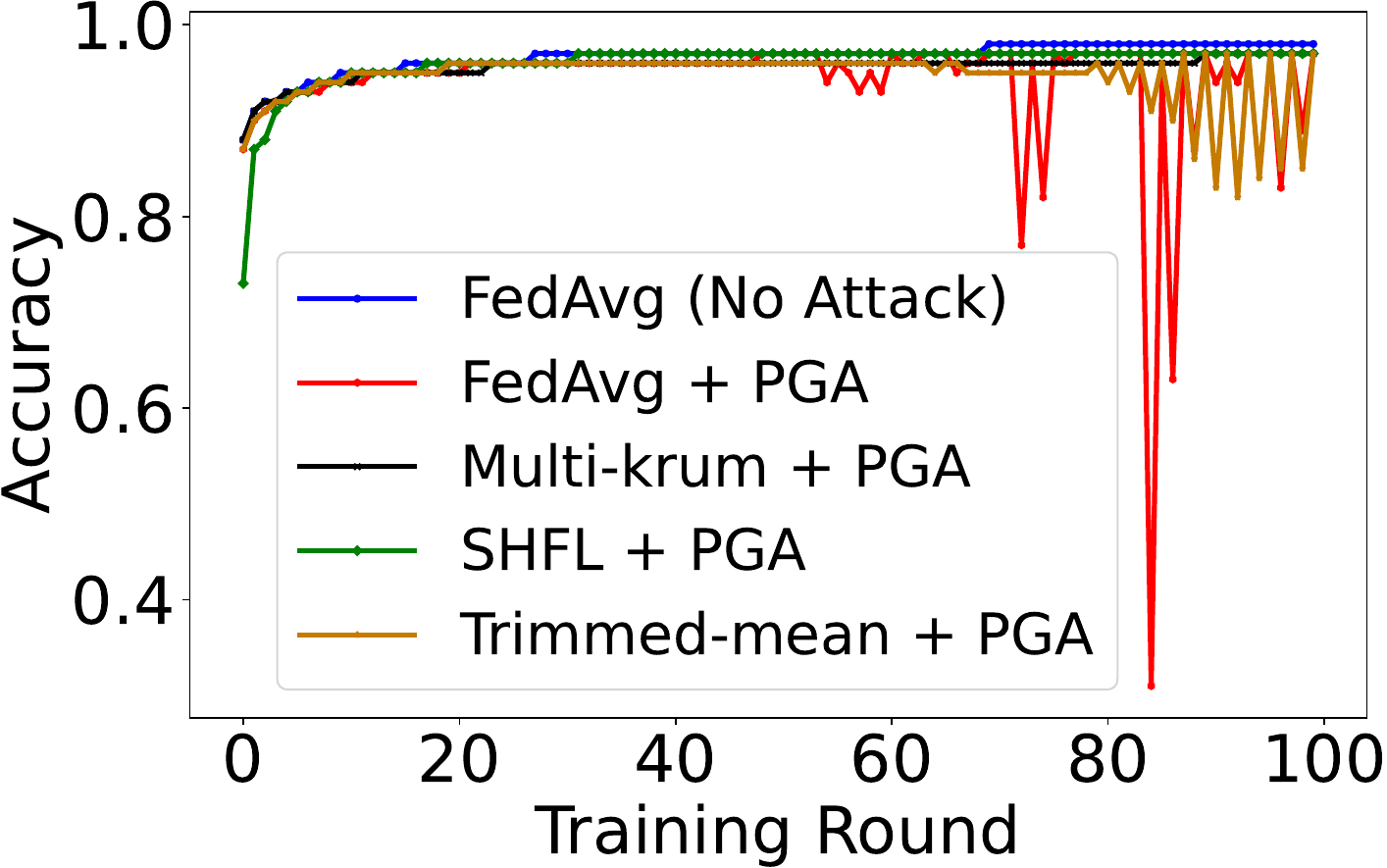}%
      \label{fig:mnist-iid-lf30}}\hspace{2 mm}
   \subfloat[\normalfont{IID (40 attackers)}]{%
      \includegraphics[width= 40 mm, height=35 mm]{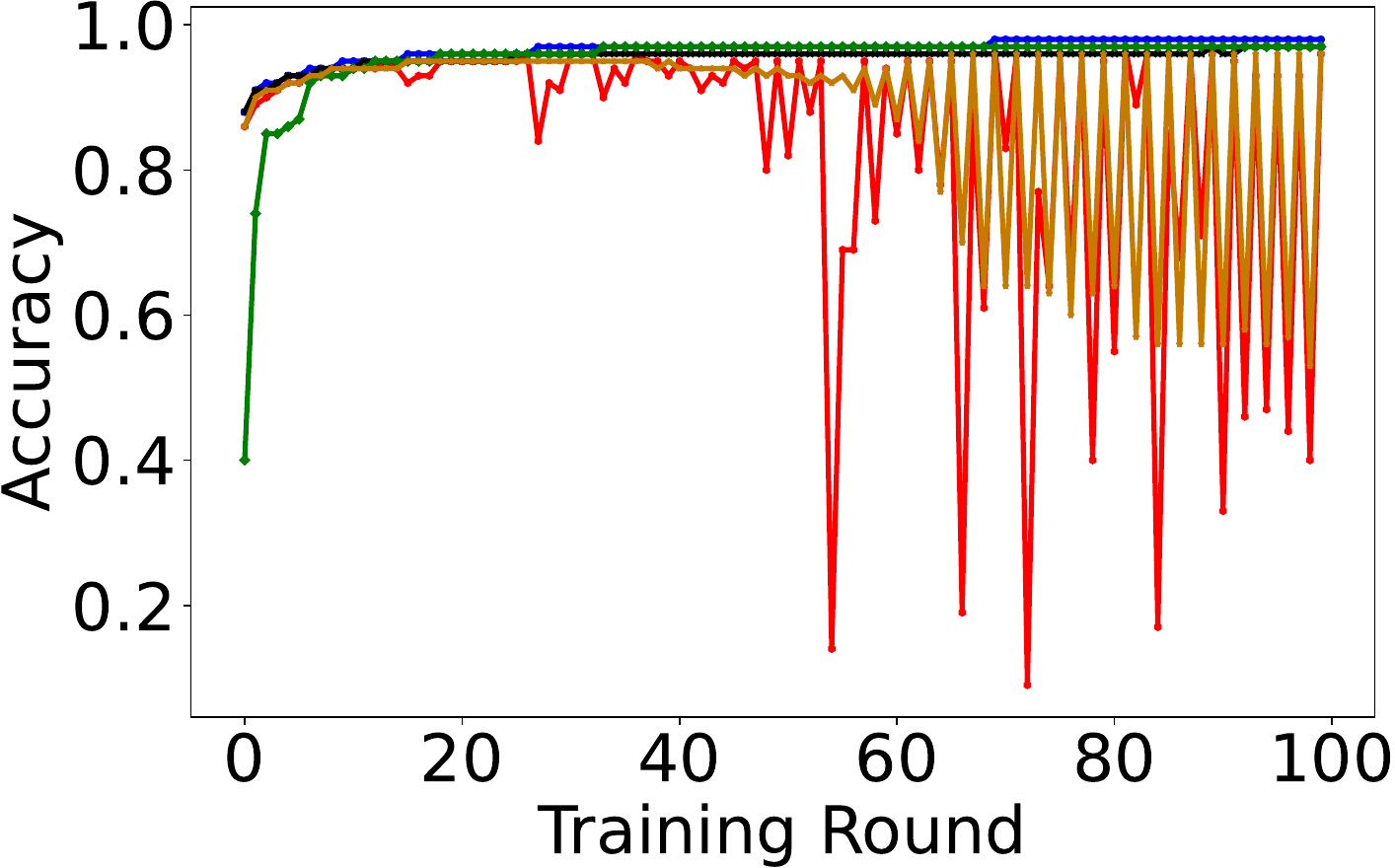}%
      \label{fig:mnist-iid-lf40}}\hspace{2 mm}
   \subfloat[\normalfont{non-IID (30 attackers)}]{%
      \includegraphics[width= 40 mm, height=35 mm]{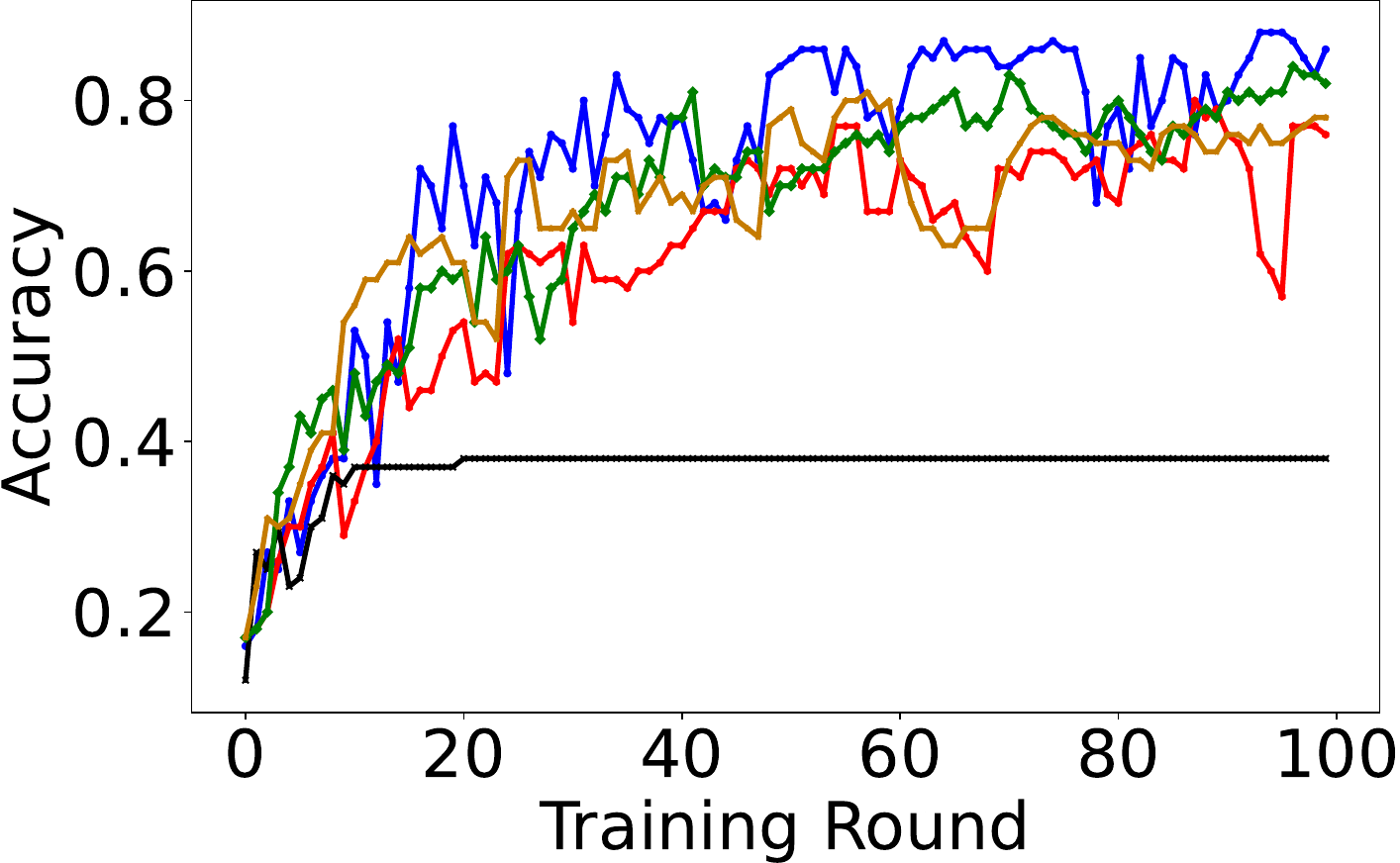}%
      \label{fig:mnist-noniid-lf30}}\hspace{2 mm}
   \subfloat[\normalfont{non-IID (40 attackers)}]{%
      \includegraphics[width= 40 mm, height=35 mm]{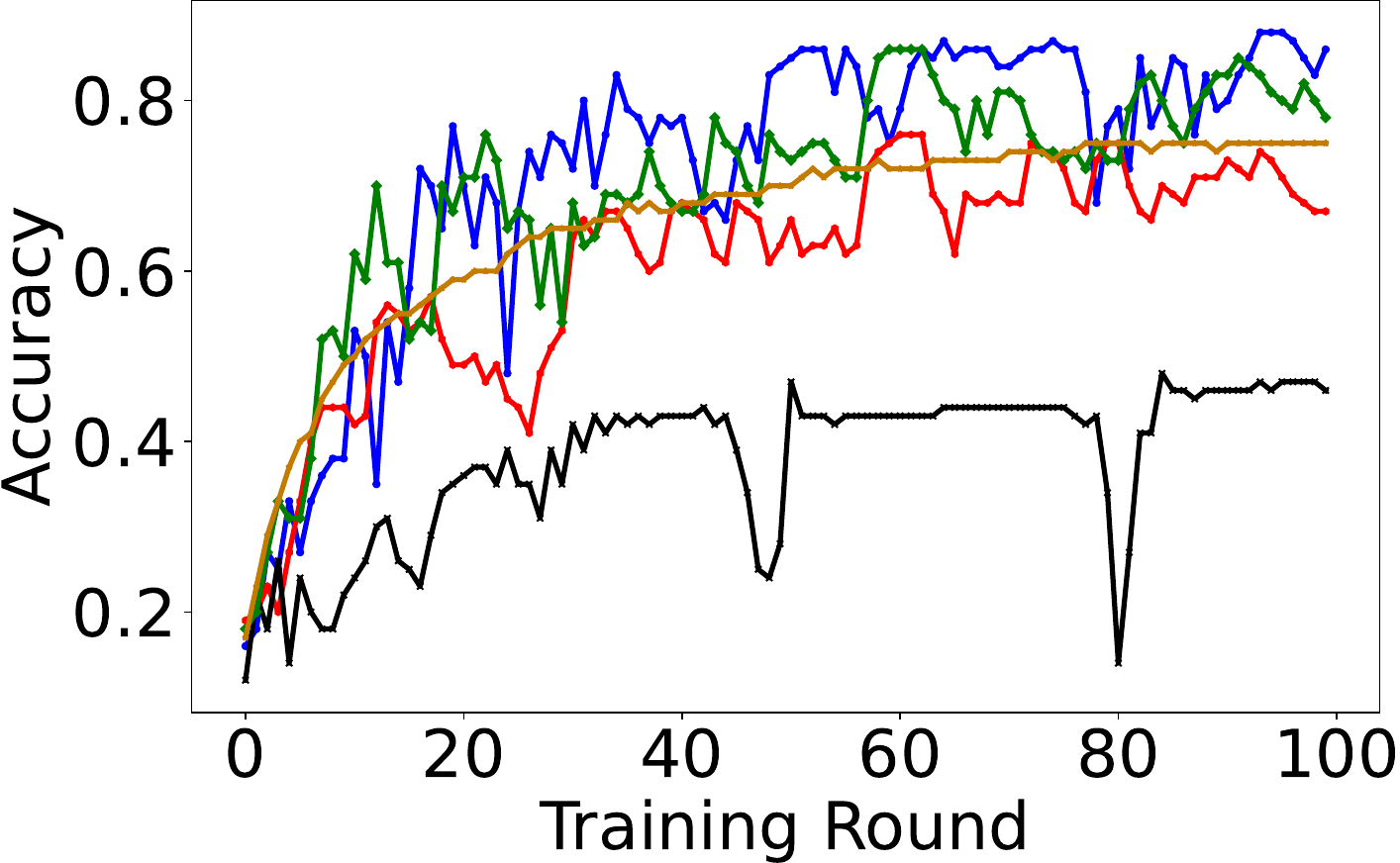}%
      \label{fig:mnist-noniid-lf40}}

  \caption{Accuracy of defense mechanisms using the \textbf{2NN} model on \textbf{MNIST} dataset for IID and non-IID scenarios under \textbf{LF} attack (30 and 40 attackers).}
  \label{fig:mnist-lf}\vspace{-2 mm}
\end{figure*}
\begin{figure*}[h!]
  \centering
  \subfloat[ \normalfont{IID (30 attackers)}]{%
      \includegraphics[width= 40 mm, height=35 mm]{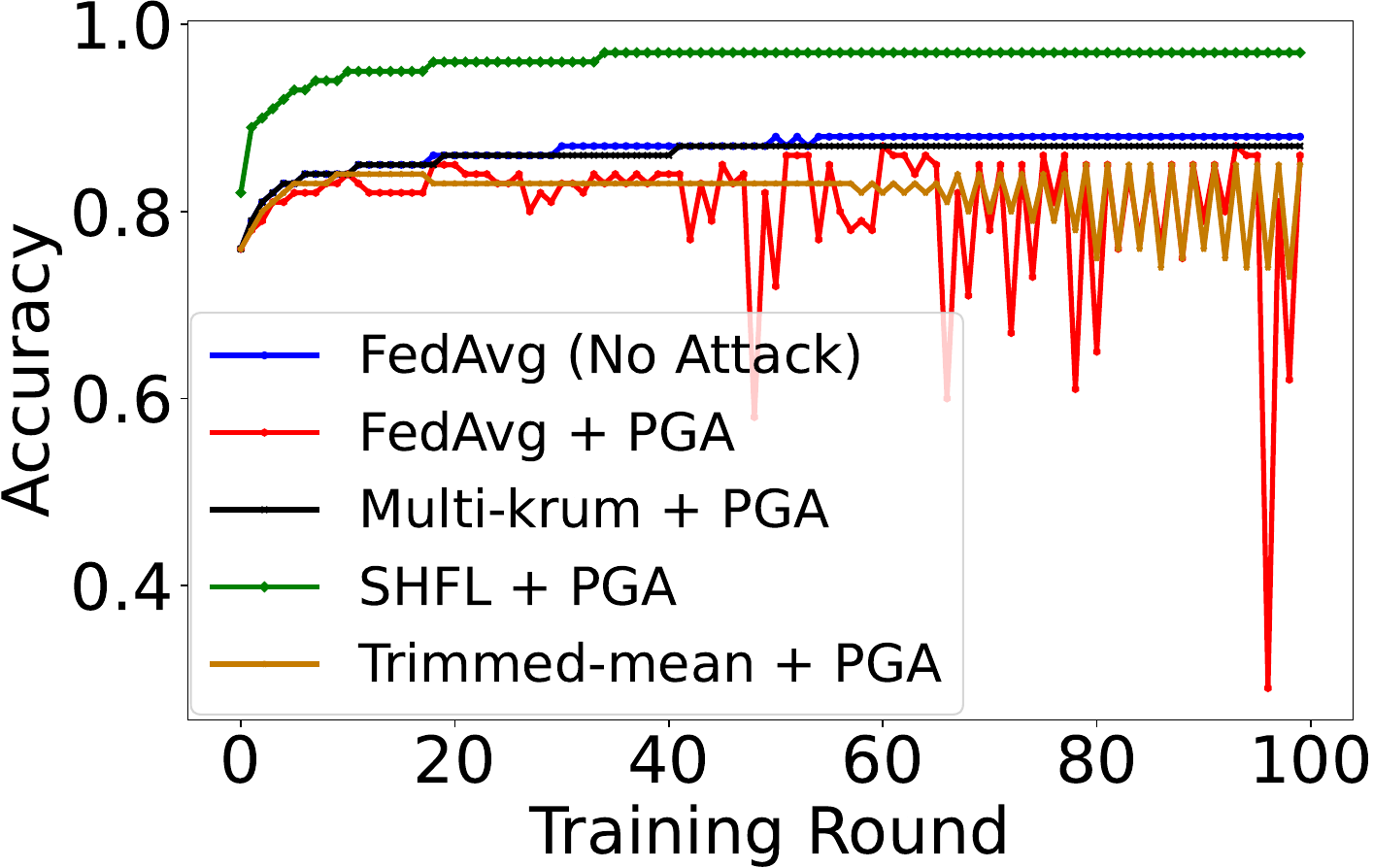}%
      \label{fig:fmnist-iid-lf30}}\hspace{2 mm}
   \subfloat[\normalfont{IID (40 attackers)}]{%
      \includegraphics[width= 40 mm, height=35 mm]{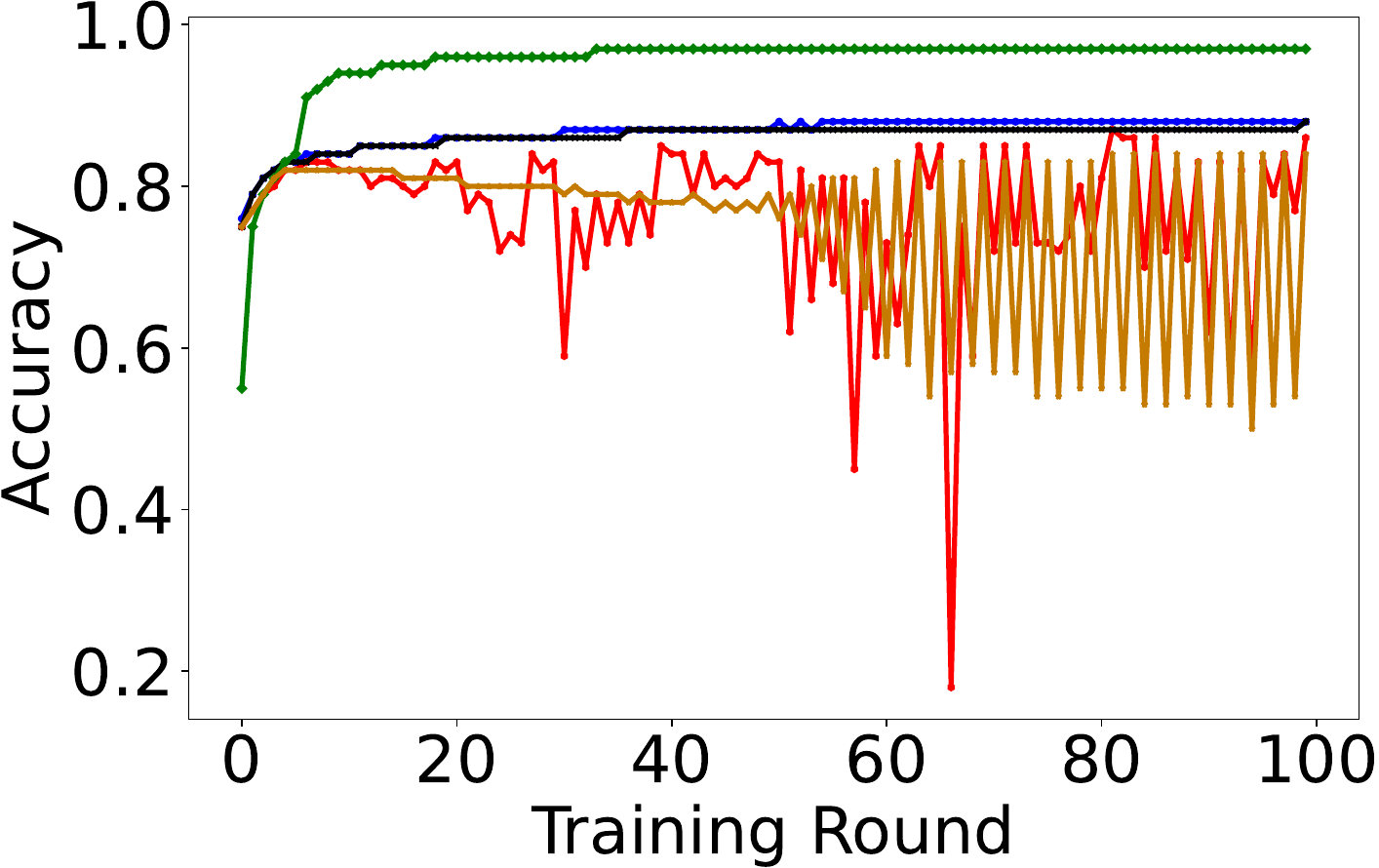}%
      \label{fig:fmnist-iid-lf40}}\hspace{2 mm}
   \subfloat[\normalfont{non-IID (30 attackers)}]{%
      \includegraphics[width= 40 mm, height=35 mm]{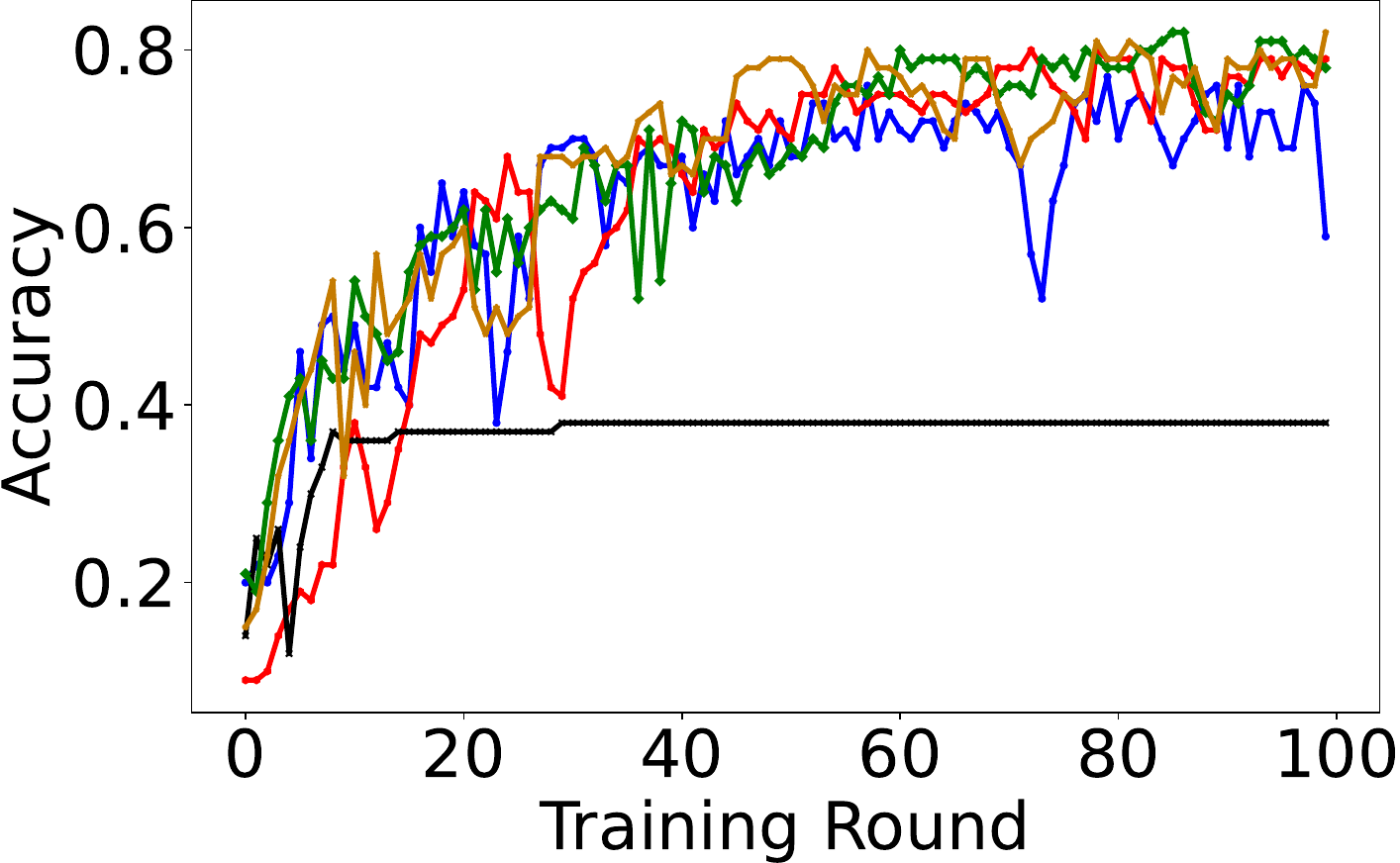}%
      \label{fig:fmnist-noniid-lf30}}\hspace{2 mm}
   \subfloat[\normalfont{non-IID (40 attackers)}]{%
      \includegraphics[width= 40 mm, height=35 mm]{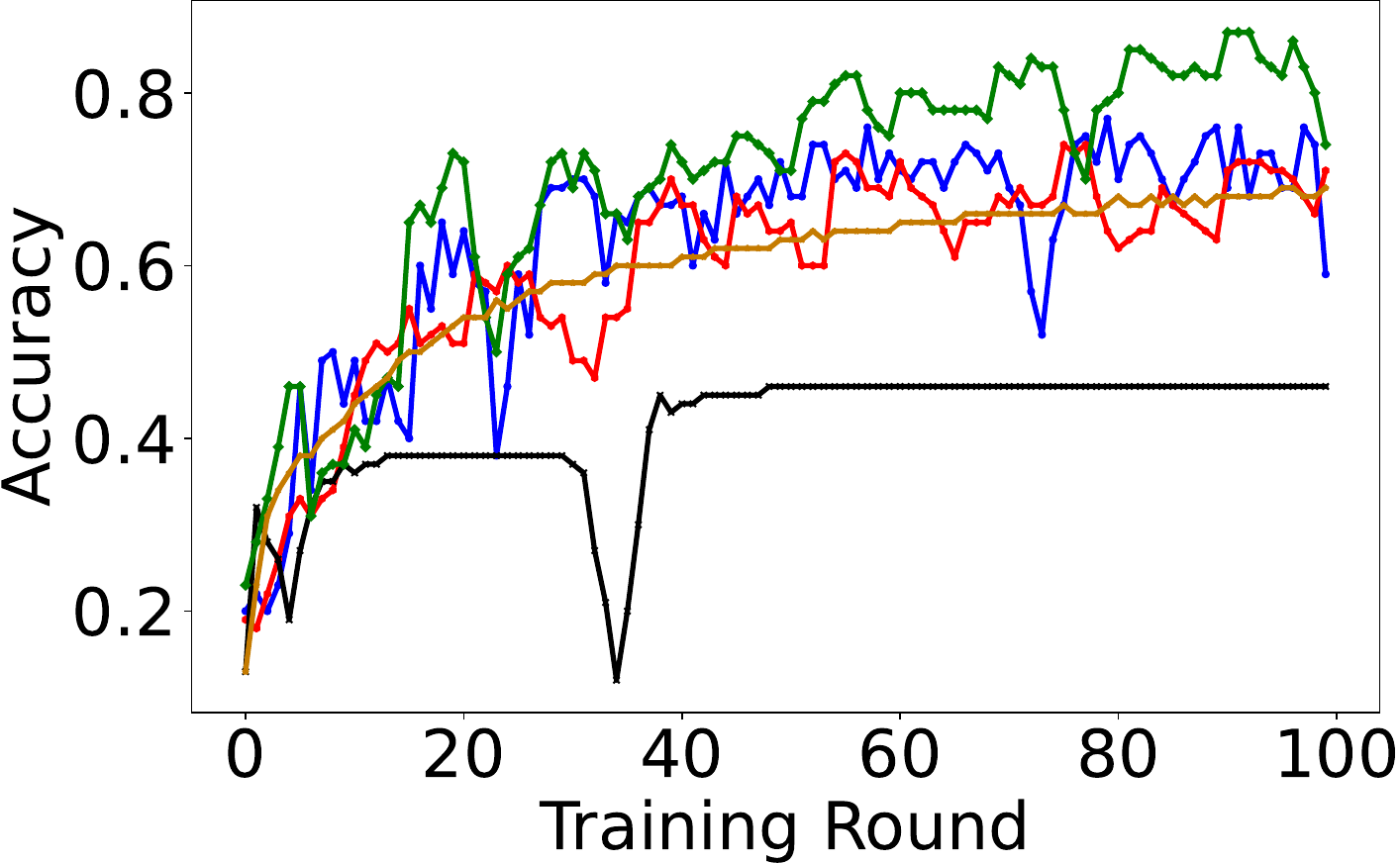}%
      \label{fig:fmnist-noniid-lf40}}

  \caption{Accuracy of defense mechanisms using the \textbf{2NN} model on \textbf{FMNIST} dataset for IID and non-IID scenarios under \textbf{LF} attack (30 and 40 attackers).}
  \label{fig:fmnist-lf}\vspace{-2 mm}
\end{figure*}
\begin{figure*}[h!]
  \centering
  \subfloat[ \normalfont{IID (30 attackers)}]{%
      \includegraphics[width= 40 mm, height=35 mm]{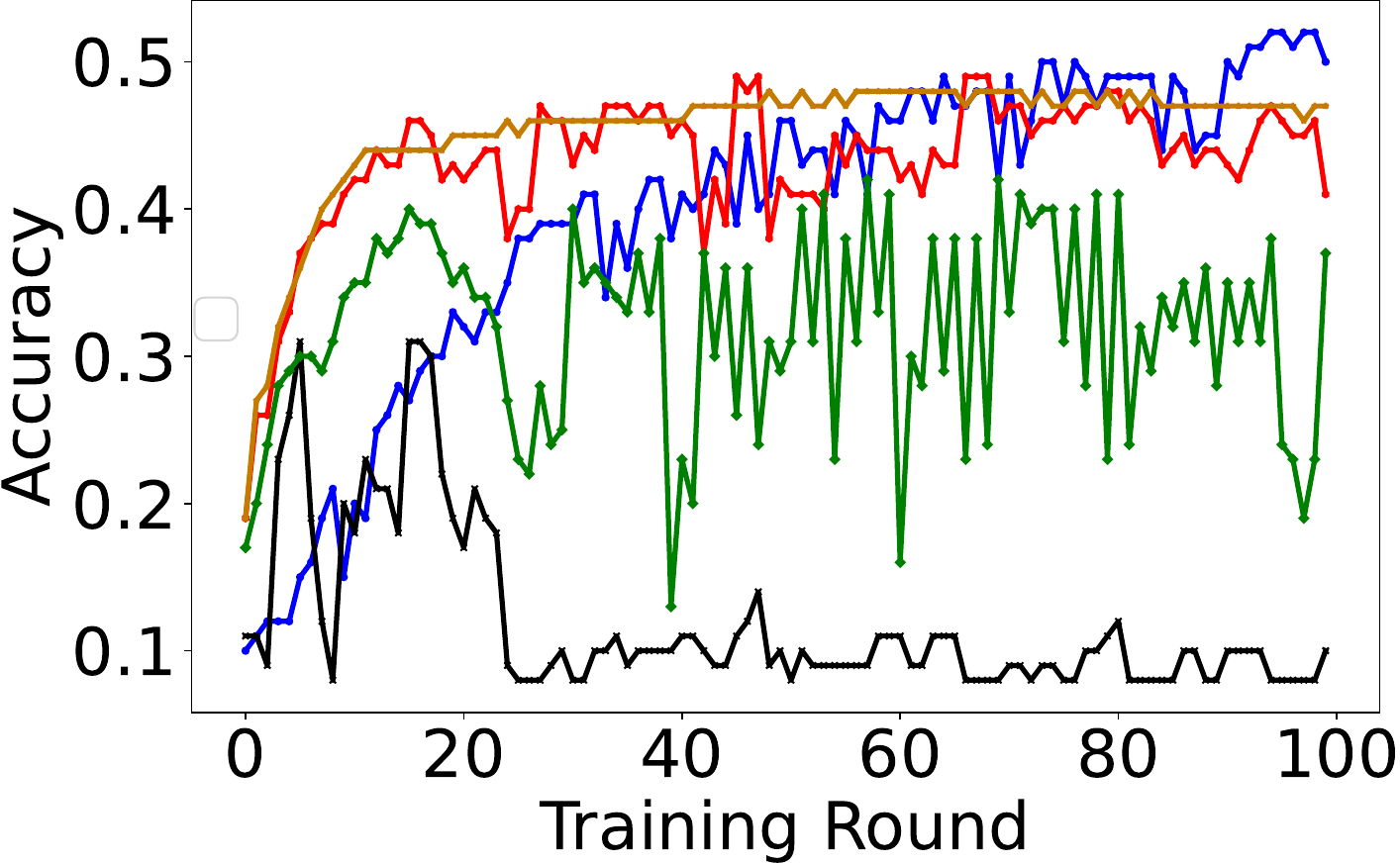}%
      \label{fig:cifar-iid-lf30}}\hspace{2 mm}
   \subfloat[\normalfont{IID (40 attackers)}]{%
      \includegraphics[width= 40 mm, height=35 mm]{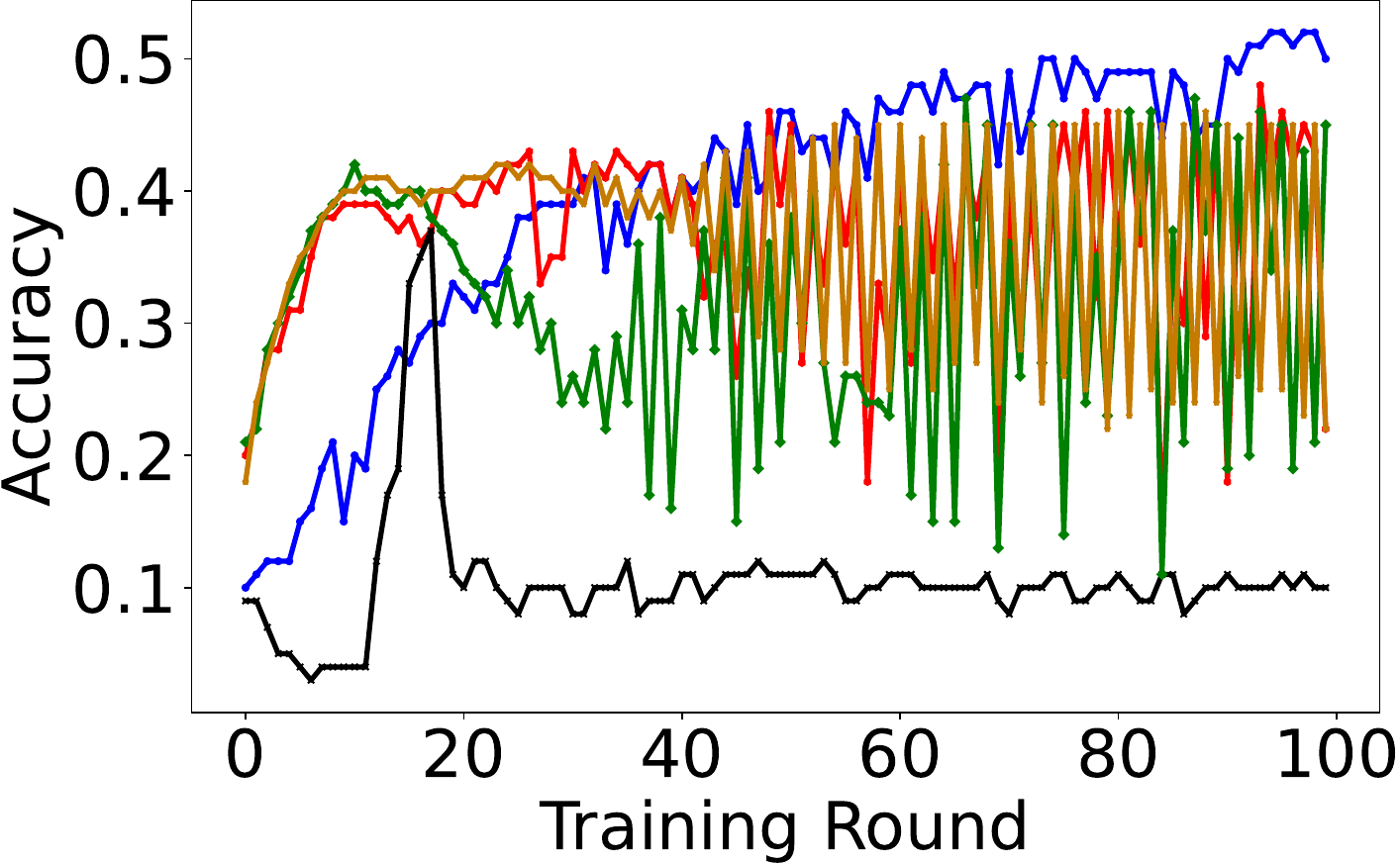}%
      \label{fig:cifar-iid-lf40}}\hspace{2 mm}
   \subfloat[\normalfont{non-IID (30 attackers)}]{%
      \includegraphics[width= 40 mm, height=35 mm]{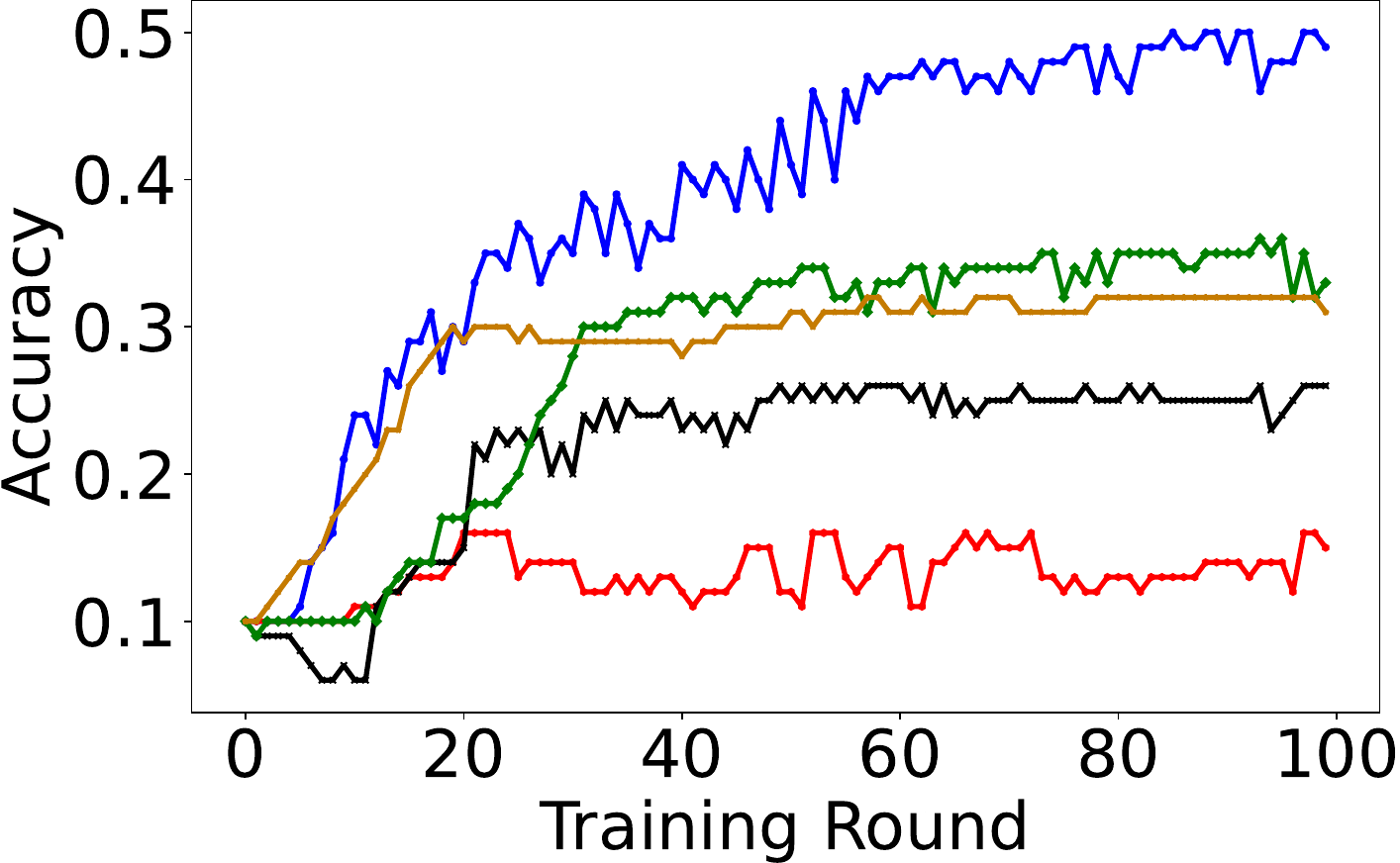}%
      \label{fig:cifar-noniid-lf30}}\hspace{2 mm}
   \subfloat[\normalfont{non-IID (40 attackers)}]{%
      \includegraphics[width= 40 mm, height=35 mm]{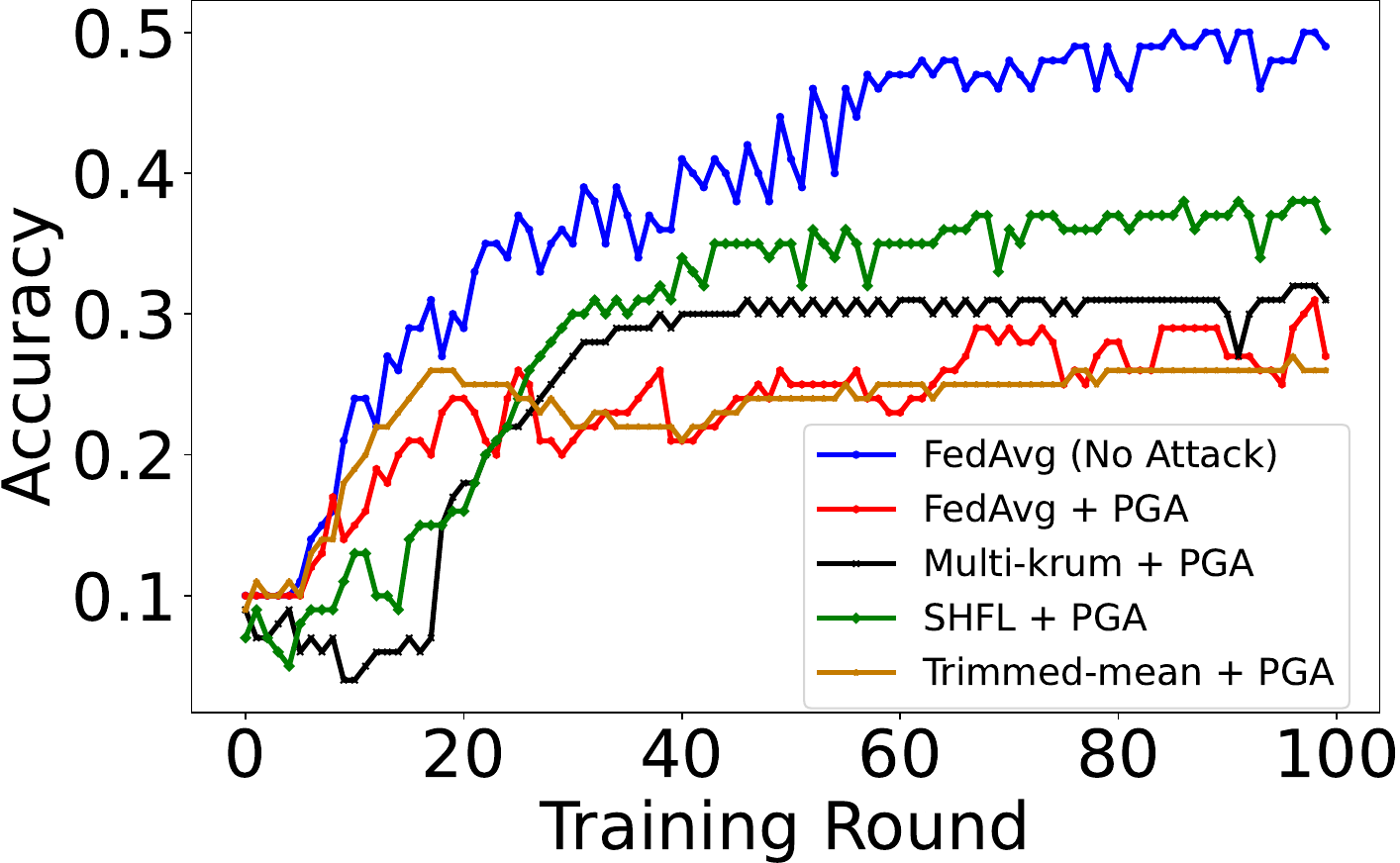}%
      \label{fig:cifar-noniid-lf40}}

  \caption{Accuracy of defense mechanisms using the \textbf{CNN} model on \textbf{CIFAR-10} dataset for IID and non-IID scenarios under \textbf{LF} attack (30 and 40 attackers).}
  \label{fig:cifar-lf}\vspace{-4 mm}
\end{figure*}

After evaluating the performance of defense mechanisms against the PGA attack, we conducted a second set of experiments to assess the resilience of defense mechanisms against the LF attack. In the IID setting for the MNIST dataset (Fig. \ref{fig:mnist-lf}), only SHFL and Multi-krum demonstrate stable performance. Both FedAvg and Trimmed-mean frequently select attacker nodes, leading to significant drops in accuracy. However, all defense mechanisms are able to mitigate the impact of attackers on the global model, achieving maximum accuracy levels exceeding 95\% (Table \ref{table:tracelf}). For non-IID data, Multi-krum's fairness problem limits its maximum accuracy to less than 50\%. This issue arises because Multi-krum selects the top $k$ clients from $n$ based solely on the average Euclidean distance between their local models and those of IoT nodes in the same edge network. To address this issue, SHFL first identifies the top $n-a$ models (where $a$ is the estimated number of attackers per edge) and then randomly selects $k$ clients from this pool. This approach not only filters out nodes with significantly divergent models but also ensures all trusted nodes have an equal opportunity to participate in training. As a result, SHFL achieves an accuracy of 86\% by incorporating data from all trusted nodes to build the edge-aggregated model that significantly improves fairness and performance in non-IID settings.

For the FMNIST dataset (which is more complex than MNIST), SHFL achieves at least 10\% higher accuracy compared to other defense methods (Fig. \ref{fig:fmnist-lf}). Trimmed-mean shows the worst performance among defense mechanisms (Trimmed-mean, Multi-krum, and SHFL). It is designed to discard extreme values, assuming they are more likely to be malicious. However, in the IID setting, the updates are generally more similar, and an LF attack may not produce extreme outliers, but instead produce values that are close enough to the benign updates to avoid being filtered. This causes Trimmed-mean to select malicious updates for the AGR process, leading to a significant drop in accuracy. For 30 attackers in the non-IID setting (Fig. \ref{fig:fmnist-noniid-lf30}), except Multi-krum, all defense mechanisms show similar performance. However, by increasing the number of attackers to 40, SHFL outperforms other methods considerably. As shown in Fig. \ref{fig:fmnist-noniid-lf40}, the maximum accuracy of FedAvg is higher than that for Trimmed-mean. This is caused by the fact that in non-IID scenarios, trained models of different clients can vary significantly due to the diversity in local data distributions. As a result, Trimmed-mean may fail to filter out malicious updates, as the benign updates themselves are already diverse. Besides, Trimmed-mean removes a fixed number of extreme values in each dimension, but in non-IID settings, some benign updates may naturally deviate from the global model due to data heterogeneity. This reduces the defense's effectiveness in distinguishing between benign and adversarial updates. 

Fig. \ref{fig:cifar-lf} presents the evaluation results for the CIFAR-10 dataset. In the IID setting with 30 LF attackers (Fig. \ref{fig:cifar-iid-lf30}), Trimmed-mean outperforms SHFL and Multi-krum. Notably, this was the only case in our experiments where SHFL did not achieve the highest accuracy within 100 rounds. The reason behind this is that, in CIFAR-10, the presence of 30 LF attackers has a relatively weak impact on the global model and by simply ignoring them during client selection, the defense mechanisms can achieve better performance compared to removing them. This occurs as filtering out malicious clients may also inadvertently exclude some benign clients whose data distribution differs from that of the majority, leading to their exclusion from the AGR process. This is further demonstrated by FedAvg, which randomly selects clients and achieves results similar to those in a non-adversarial network. However, when the number of attackers increases to 40, SHFL outperforms Trimmed-mean in terms of the maximum accuracy. In the non-IID setting, the impact of the LF attack on the global model is more pronounced. As a result, SHFL consistently achieves higher accuracy after 30 rounds by detecting and filtering malicious updates. By increasing the number of attackers to 40 (Fig. \ref{fig:cifar-noniid-lf40}), Trimmed-mean shows the worst performance, further supporting the idea that Trimmed-mean fails to account for the diversity of benign updates in the non-IID setting when filtering outliers. Table \ref{table:tracelf} shows the maximum accuracy of defense methods and FedAvg against the LF attack in 100 training rounds. 

\begin{table*}[htbp]
\centering
\caption{The maximum accuracy of different defense mechanisms and FedAvg in the presence of 30 and 40 LF attackers in 100 training rounds.}\vspace{-2 mm}
{ 
\begin{tabular}{|c|c|c|c|c|c|c|c|}
\hline
\multirow{3}{*}{Defense} & \multicolumn{2}{c|}{MNIST (2NN)} & \multicolumn{2}{c|}{FMNIST (2NN)} & \multicolumn{2}{c|}{CIFAR-10 (CNN)} \\ \cline{2-7} 
                                  & IID      & non-IID   & IID    & non-IID & IID     & non-IID \\ \hline
FedAvg (No Attack)                & 0.97              & 0.88              & 0.88            & 0.76            & 0.52             & 0.50             \\ \hline
FedAvg (30)                  & 0.97              & 0.79              & 0.87            & 0.80            & 0.49             & 0.16             \\ \hline
FedAvg (40)                  & 0.96              & 0.75              & 0.86            & 0.73            & 0.48             & 0.31             \\ \hline
Multi-krum (30)                & 0.96              & 0.38              & 0.87            & 0.38            & 0.31             & 0.25             \\ \hline
Multi-krum (40)                & 0.96              & 0.47              & 0.87            & 0.46            & 0.36             & 0.31             \\ \hline
SHFL (30)                & 0.97              & 0.83              & 0.97            & 0.82            & 0.41             & 0.35             \\ \hline
SHFL (40)                & 0.97              & 0.86              & 0.97            & 0.87            & 0.46             & 0.38             \\ \hline
Trimmed-mean (30)        & 0.97              & 0.80              & 0.85            & 0.81            & 0.48             & 0.32             \\ \hline
Trimmed-mean (40)        & 0.96              & 0.75              & 0.84            & 0.69            & 0.45             & 0.26             \\ \hline
\end{tabular}\label{table:tracelf}
} 
\end{table*}

\section{conclusion}
\vspace{2 mm}
In this paper, we presented SHFL, a secure hierarchical FL framework designed for edge networks to make them resilient against poisoning attacks. SHFL leverages the hierarchical structure of edge networks and introduces a defense mechanism against model/data poisoning attacks by proposing 1) a client selection algorithm at the edge to filter out poisoned updates and 2) an AGR process designed based on convex optimization theory to reduce the impact of poisoned edge models on creating the global model. Through an extensive set of experiments, we examined the performance of SHFL in a network with 100 IoT nodes and 10 edge servers by considering various data distributions, datasets, and neural network models. The evaluation results revealed that, compared to state-of-the-art methods, SHFL offers strong resilience and achieves much higher maximum accuracy in 100 training rounds, particularly in the non-IID setting.
\section{ACKNOWLEDGMENT}
This work was funded by the National Intelligence and Security Discovery Research Grant (NI220100111), Australia.
\bibliographystyle{IEEEtran}
\bibliography{shfl}

\begin{IEEEbiography}
    [{\includegraphics[height=35 mm, width=25 mm]{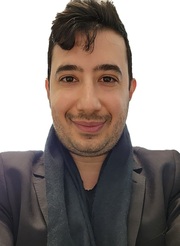}}]{Omid Tavallaie}
received his PhD degree from the School of Computer Science, the University of Sydney, Sydney, Australia, in 2021. He is a postdoctoral researcher and lecturer at the School of Computer Science, the University of Sydney, and a member of the Centre for Distributed and High-Performance Computing. His research interests include designing optimized protocols and solutions for wireless and IoT networks and solving complex problems for resource-limited devices.
\end{IEEEbiography}

\begin{IEEEbiography}
    [{\includegraphics[height=35 mm, width=25 mm]{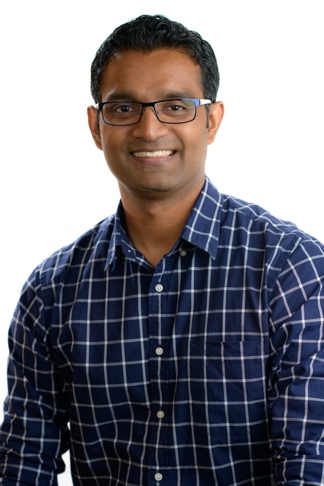}}]{Kanchana Thilakarathna }
is a Senior Lecturer at the School of Computer Science, The University of Sydney. He was a visiting academic at the Electrical and Computer Engineering at the University of Maryland. He received his PhD from University of New South Wales in 2015 with the Malcolm Chaikin Prize, awarded to the Best Engineering PhD Thesis, and then worked as a Research Scientist at the Information Security and Privacy research group at Data61-CSIRO, Australia. His research interests include cybersecurity, network security, data privacy, mobile and distributed computing. He is a recipient of many research grants from industry and government organizations such as the Defense Science and Technology Organization, Defense Innovation Network, Data61-CSIRO, and Office of National Intelligence, which have been recognized by the Dean’s Award for Industry Collaborations, Google Faculty Awards and Facebook/Meta Research Awards. 
\end{IEEEbiography}

\begin{IEEEbiography}
    [{\includegraphics[height=35 mm, width=25 mm]{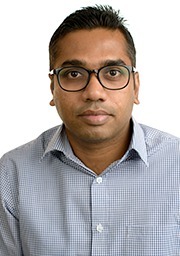}}]{Suranga Seneviratne }
    is a Senior Lecturer in Security at the School of Computer Science, The University of Sydney. He received his Ph.D. from the University of New South Wales, Australia in 2015. His current research interests include privacy and security in mobile systems, AI applications in security, and behavior biometrics. Before moving into research, he worked nearly six years in the telecommunications industry in core network planning and operations. He received his bachelor degree from University of Moratuwa, Sri Lanka in 2005.
\end{IEEEbiography}
\begin{IEEEbiography}
    [{\includegraphics[height=35 mm, width=25 mm]{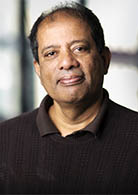}}]{Aruna Seneviratne}
(Senior Member, IEEE) is currently a foundation professor of telecommunications with the University of New South Wales, Sydney, Australia, where he holds the Mahanakorn chair of telecommunications. He was with a number of other universities in Australia, UK, and France, as well as industrial organizations, including Muirhead, Standard Telecommunication Labs, Avaya Labs, and Telecom Australia (Telstra). He held visiting appointments with INRIA, France. His research interests include physical analytics, technologies that enable applications to interact intelligently and securely with their environment in real time. Recently, his team has been working on using these technologies in behavioural biometrics, optimizing the performance of wearables, and IoT system verification. He was the recipient of several fellowships, including one at British Telecom and one at Telecom Australia Research Labs.
\end{IEEEbiography}
\begin{IEEEbiography}
    [{\includegraphics[height=35 mm, width=25 mm]{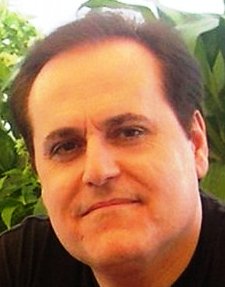}}]{Albert Y. Zomaya }
is the Peter Nicol Russell Chair Professor of Computer Science in the School of Computer Science, Sydney University, and serves as the Director of the Centre for Distributed and High-Performance Computing. Professor Zomaya has published over 800 scientific papers and articles and is the author, co-author, or editor of more than 30 books. He is the past Editor in Chief of the IEEE Transactions on Computers, the IEEE Transactions on Sustainable Computing, and the ACM Computing Surveys.
Professor Zomaya received numerous accolades, including Fellowships of the IEEE, AAAS, and the IET. Also, he is a Fellow of the Australian Academy of Science, a Fellow of the Royal Society of New South Wales, a Foreign Member of Academia Europaea, and a Member of the European Academy of Sciences and Arts. His research interests are in parallel and distributed computing, networking, and complex systems.
\end{IEEEbiography}

\end{document}